\pdfoutput=1 

\documentclass{cta-author}

\usepackage{threeparttable}  
\usepackage{array}

\newcolumntype{L}[1]{>{\raggedright\let\newline\\\arraybackslash\hspace{0pt}}p{#1}}
\newcolumntype{C}[1]{>{\centering\let\newline\\\arraybackslash\hspace{0pt}}p{#1}}
\newcolumntype{R}[1]{>{\raggedleft\let\newline\\\arraybackslash\hspace{0pt}}p{#1}}

\usepackage{url}
\usepackage{multirow}
\usepackage{subcaption} % for subfigure
\begin{document}

\title{Deep convolutional neural networks for face and iris presentation attack detection: Survey and case study} %. 
\author{\au{Yomna Safaa El-Din$^{1\corr}$}, \au{Mohamed N. Moustafa$^{2}$}, \au{Hani Mahdi$^{1}$}}

\address{\add{1}{Department of Computer and Systems Engineering, Ain Shams University,  Cairo, Egypt}
\add{2}{Department of Computer Science and Engineering, The American University in Cairo, New Cairo, Egypt}
\email{yomna.safaa-eldin@eng.asu.edu.eg}
}

\begin{abstract}
	% for IET max 200 words
	Biometric presentation attack detection is gaining increasing attention. Users of mobile devices find it more convenient to unlock their smart applications with finger, face or iris recognition instead of passwords. In this paper, we survey the approaches presented in the recent literature to detect face and iris presentation attacks. Specifically, we investigate the effectiveness of fine tuning very deep convolutional neural networks to the task of face and iris antispoofing. We compare two different fine tuning approaches on six publicly available benchmark datasets. Results show the effectiveness of these deep models in learning discriminative features that can tell apart real from fake biometric images with very low error rate. Cross-dataset evaluation on face PAD showed better generalization than state of the art. We also performed cross-dataset testing on iris PAD datasets in terms of equal error rate which was not reported in literature before. Additionally, we propose the use of a single deep network trained to detect both face and iris attacks. We have not noticed accuracy degradation compared to networks trained for only one biometric separately. Finally, we analyzed the learned features by the network, in correlation with the image frequency components, to justify its prediction decision.

\end{abstract}

\maketitle

\section{Introduction}\label{sec1}

Biometric recognition has been increasingly used in recent application that need authentication and verification. Instead of normal username and password or token-based authentication, the use of biometric traits like face, iris or fingerprints are more convenient to the users and so has gained a lot of popularity specially after the wide spread of smartphones and its usage in a lot of areas, such as payment. 
However, with the increase of technology, spoofing these biometric traits has become more easy. For example, an attacker can use a photograph or a video replay of the face or iris or a person to gain access to his/her smartphone or any application that needs authorization. Such act is referred to by the ISO/IEC 30107-1:2016 standard as presentation attack (PA) and the biometric or object used in PA is called an artifact or presentation attack instrument (PAI). This has lead to the increase of interest in designing algorithms that guard against these attacks, and so are called presentation attack detection (PAD) algorithms.

Lots of research has been done in this area, starting with methods that basically depend on designing and extracting hand-crafted features from the acquired images then using conventional classifiers to detect bona-fide (real) from attack presentations. Literature shows that methods relying on manually engineered features are suitable for solving the PAD problem for face and iris recognition systems. However, the design and selection of handcrafted feature extractors is mainly based on expert knowledge of researchers on the problem. Consequently, these features often reflect limited aspects of the problem and are often sensitive to varying acquisition conditions, such as camera devices, lighting conditions and PAIs. This causes their detection accuracy to vary significantly among different databases, indicating that the handcrafted features have poor generalizability and so do not completely solve the PAD problem.

In recent years, deep learning has evolved and the use of deep neural networks or convolutional neural networks (CNN) has proven to be effective in many computer vision tasks especially with the availability of new advanced hardware and large data. CNNs have been successfully used for vision problems like image classification and object detection. This has encouraged many researchers to incorporate deep learning in the PAD problem, and rely on a deep network or CNN to learn features that discriminate between bona-fide and attack face~\cite{319_18_f, 317_18_f} or iris~\cite{305_18_i, 306_18_i} or both~\cite{312_18_fi}, instead of using hand-crafted features. Several of these researches used CNN only for feature extraction followed by a classic classifier like SVM, others designed custom networks or combined information from CNN and hand-crafted features. 

Several surveys for presentation attack detections in either face or iris recognition are available in literature. Czajka and Bowyer provided a thorough assessment of the state-of-the-art for iris PAD in~\cite{183_18} and concluded that PAD for iris recognition is not a solved problem yet. Their main focus was on iris presentation attack categories, their countermeasures, competitions and applications.

For PAD in face recognition systems, Raghavendra and Bush provided a comprehensive survey in~\cite{526_17_f} describing different types of presentation attack and face artifacts, and showing the vulnerability of commercial face recognition systems to presentation attack. They survey and evaluate fourteen state-of-the-art face PAD algorithms on CASIA face-spoofing database and discuss the remaining challenges and open issues for a robust face presentation attack detection system. In~\cite{503_17_f}, Ghaffar and Mohd focused their review on face PAD systems on smartphones with recent face datasets captured with mobile devices.

Later, Li~\etal~\cite{502_18_f} provided a comprehensive review of more recent PAD algorithms in face recognition, discussing the available datasets and reported results. They proposed a color-LBP based countermeasure as a case study and evaluated it on two benchmark face datasets highlighting the importance of cross-dataset testing.

In this work, we survey most of the recent PAD approaches proposed in literature for both face and iris recognition applications, along with the competitions and benchmark datasets. Then for a case study, we assess the performance of recent well-known deep CNN architectures on the task of face and iris presentation attack detection.

For assessment we use 3 different face datasets and 3 iris datasets. 
Using just the RGB images as input, we finetune the deep network's weights then perform intra and cross-dataset evaluation to discuss the generalization ability of these deep models. The experiments are first done with separate models trained for each of face and iris biometric, then we experiment with training a single network to generally tell apart bona-fide from attack samples whether the sample is a face or an iris image. Such generic PAD network could perfectly differentiate between the real and attack images. The specific features learned by this network are analyzed showing that these features are highly related to high frequency regions in the input images. We finally use gradient-weighted class activation maps~\cite{gradcam} to visualize what the network focuses on when deciding if a sample is bona-fide or not.

Hence, the main contributions of this work include: (1) Survey the recent methods for presentation attack detection in both face and iris modalities with competitions and datasets, (2) Demonstrate the effectiveness of CNN for the PAD problem, by finetuning and comparing recent deep CNN architectures on 3 benchmark face datasets and 3 benchmark iris datasets, experimenting with two finetuning approaches and applying cross-dataset evaluation, (3) Train a single network for presentation attack detection of the two biometric modalities, and compare the results with networks trained on single biometric, (4) Analyze specific features learned by the commonly-trained network and (5) Finally, visualize the class activation maps of the trained network on sample bona-fide and attack images.

The rest of the paper is organized as follow, we first review the PAD approaches which can broadly be categorized into active (Section~\ref{sec:active}) vs passive (Section~\ref{sec:passive}) PAD techniques. Then in Section~\ref{sec:compet_db}, we summarize the face and iris PAD competitions and benchmark databases, followed by Section~\ref{sec:prop} where we explain the proposed PAD approach. Experiments, results and analysis are presented in Section~\ref{sec:exp} and finally conclusion and future work in Section~\ref{sec:conc}.

%-------------------------------------------------------------------------

\section{Active PAD}\label{sec:active}
Active Presentation Attack Detection can either be hardware-based or depending on challenge-response, both will be highlighted in details in this section.

\subsection{Hardware-based}
Such methods depend on capturing the automatic response of the biometric trait (face/iris) to a certain stimulus. For example in iris liveness detection, the use of eye hippus, which is the permanent oscillation that the eye pupil presents even under uniform lighting conditions, or the dilation of pupil in reaction to sudden illumination changes
~\cite{111_04,104_12,107_06,150_15,172_14}. Variations of pupil dilation were calculated by Bodade~\etal~\cite{148_09} from multiple iris images while Huang~\etal~\cite{149_13} used pupil constriction to detect iris liveness. Other researchers used the iris image brightness variations after light stimuli in some predefined iris regions~\cite{119_07}. Or the reflections of infrared light on the moist cornea when stimulated with light sources positioned randomly in space.
Lee~\etal~\cite{145_10} used additional infrared sensors during iris image acquisition to construct the full 3d shape of the iris to aid in the detection of fake iris images or contact lenses. These approaches rely on an external specific device to be added to the sensor/camera, hence the naming "hardware-based techniques". However, not all hardware-based methods are active, some such approaches detect facial thermogram~\cite{hwp1_15_f, hwp2_15_f}, blood pressure, fingerprint sweat, gait etc passively in a non-intrusive manner.

\subsection{Challenge-response}
In these methods, the user is required to respond to a "challenge" instructed by the system. The system requests the user to perform some specific movement like "blink right eye" or "rotate the head clockwise"~\cite{531_11_f,562_15_f,528_08_f} or gaze towards a predefined stimulus~\cite{563_13_f}. 

\subsection{Drawbacks of active PAD methods}
Although active spoof detection methods may provide higher presentation attack detection rates and robustness against different presentation attack types, they are more expensive and less convenient to users than the software-based approaches which do not include any additional devices beside the standard camera. For the rest of this survey we will focus on software-based methods; they are more generic, less expensive, less intrusive and can easily be incorporated in real-world applications and smartphones. Table~\ref{tab:PADSum} summarizes the different approaches used in literature for detection of presentation attacks.

%-------------------------------------------------------------------------
\begin{table*}[t!]

\fwprocesstable{PAD methods in literature. \label{tab:PADSum}}
{\begin{tabular*}{\textwidth}{L{2cm}L{5.2cm}L{5cm}L{5cm}@{}}

		\toprule

PAD Method &  & Advantage & Weakness \\
\midrule

\multicolumn{2}{l}{Active (Hardware-based or Challenge-response)} &  &  \\[0.2cm]
 & \begin{tabular}[t]{@{}p{5.2cm}@{}}- Eye hippus or Pupil-dilation:~\cite{111_04,104_12,107_06,150_15,172_14,148_09,149_13,119_07,145_10}  \\ - Challenge: "blink eye" or "rotate head": ~\cite{531_11_f,562_15_f,528_08_f,563_13_f} \end{tabular} & \begin{tabular}[t]{@{}l@{}}Robust against different PA types\\ Higher PA detection rates\end{tabular} & \begin{tabular}[t]{@{}l@{}}More expensive\\ Less convenient to users \\ Rely on an external device\end{tabular} \\

\midrule

\multicolumn{2}{l}{Passive (Software-based):} &  &  \\[0.2cm]

Liveness-detection 
& \begin{tabular}[t]{@{}p{5.2cm}@{}}- Eye blinking: ~\cite{529_09_f, 528_08_f, 513_13_f, 506_07_f, 318_16_f}\\ - Facial expression changes, head and/or mouth movements: ~\cite{521_12_f,520_08_f,519_07_f}\\ - Eye movements and gaze estimationf for iris:~\cite{152_13, 174_15, 175_14, 178_15}\end{tabular} 
& \begin{tabular}[t]{@{}p{5cm}@{}}Utilize features extracted from consecutive frames.\\ Detect signs of life in captured biometric data.\\ \\\end{tabular} & 
Can be easily fooled by a replay video or by adding liveness properties to a still face image by cutting the eye/mouth parts 
\\[1cm]

Recapturing-detection & \begin{tabular}[t]{@{}p{5.2cm}@{}}- Detect abnormal movements in the video due to a hand holding the presenting media~\cite{523_16_f, 507_10_f}\\ - Track the relative movements between different facial parts~\cite{508_09_f, 509_09_f}\\ - Motion magnification~\cite{511_16_f, 513_13_f}\\ - Visual rhythm analysis~\cite{535_12_f}. \\ - Contextual analysis of~\cite{531_11_f,530_13_f,521_12_f,varF_13_f}\\ - 3d reconstruction~\cite{thrD_13_f}\end{tabular} & Very effective for replay attacks & Fail if the attack is done using an artifact like a mask\\[2cm]

Texture-based (Hand-crafted features) 
& \begin{tabular}[t]{@{}p{5.2cm}@{}}- Texture-reflectance ~\cite{538_13_f, 537_13_f, 536_13_f}\\ - Frequency ~\cite{518_04_f, 507_10_f, b_12_f,534_13_f,109_13}\\ - Image Quality analysis ~\cite{104_12, 108_09, 119_07, 155_14, 164_14,514_16_f, 515_15_f, 516_14_f,500_18_fi, 505_17_fi,104_12}\\ - Color analysis ~\cite{515_15_f, 550_15_f, 545_16_f, 546_17_f, compet_f3,518_04_f,187_16,555_18_f}.\\ - Local-descriptors~\cite{533_11_f,543_13_f,544_15_fi,552_13_f,540_13_f, 532_11_f,547_12_f,548_13_f,549_12_f,556_15_f,553_16_f,546_17_f,548_13_f,549_12_f,541_11_f,530_13_f,310_15_f,554_11_f,507_10_f,541_11_f,161_15,171_15,544_15_fi,558_14_fi}\\ - Fusion~\cite{542_11_f,541_11_f,540_13_f, 543_13_f,311_16_f,558_14_fi,305_18_i,306_18_i}\end{tabular} 
& \begin{tabular}[t]{@{}p{5cm}@{}}Detect effects caused by printers and display devicesFaster than temporal or motion-based analysis\\ Operate with only one image sample, or a selection of images, instead of video sequence.\end{tabular} 
& \begin{tabular}[t]{@{}p{5cm}@{}}Design and selection of feature extractors is based on expert knowledge.\\ Sensitive to varying acquisition conditions, such as camera devices, lighting conditions and PAIs.\\ Have poor generalizability.\end{tabular}  \\[2cm]
 
%\midrule

Learned-features 
& \begin{tabular}[t]{@{}p{5.2cm}@{}}- Extract deep features, classify classic ~\cite{315_16_f,318_16_f,304_18_f}\\ - Combine with temporal information ~\cite{311_16_f,310_15_f}\\ - Other ~\cite{301_17_f,300_18_f,302_18_f,300_18_f,313_14_f, 319_18_f, 314_17_f, 316_17_f, 317_18_f}\\ - For iris~\cite{305_18_i, 306_18_i, 307_17_i, 309_15_i,320_18_i,321_18_i}\\ - Face and iris~\cite{308_15_fi, 312_18_fi}\end{tabular} 
& \begin{tabular}[t]{@{}p{5cm}@{}}Instead of hand engineering features\\ Deep network learns features that discriminate between bona-fide and attack samples\\ Can be used jointly with hand-crafted features or motion-analysis\end{tabular} 
& \begin{tabular}[t]{@{}p{5cm}@{}}Requires more data\\ Can fail to generalize to unseen sensors\end{tabular}
\\

\botrule
\end{tabular*}}{}

\end{table*}

%-------------------------------------------------------------------------
\section{Passive Software-based PAD}\label{sec:passive}
Passive software-based antispoofing methods can be categorized into several groups. Temporal, or motion-analysis, based approaches which utilize features extracted from consecutive frames captured for the face or iris. Or texture-based methods that use single image of the biometric trait and does not depend on motion features. Such texture-based PAD methods can further be classified into either a group that depends on hand-crafted features vs. another group of recent methods that opt to learn these discriminative features.

\subsection{Liveness-detection}

The earliest category include approaches that identify whether the presented biometric is live or an artifact, such approaches are known as "liveness-detection". An artifact can be a mask or paper in case of face spoofing and paper or a glass-eye for iris. These methods depend on detecting signs of life in captured biometric data. Examples in the case of face PAD are eye blinking~\cite{529_09_f, 528_08_f, 513_13_f, 506_07_f, 318_16_f}, facial expression changes, head and/or mouth movements~\cite{521_12_f,520_08_f,519_07_f}. For the iris PAD, Komogortsev~\etal~\cite{152_13} extracted liveness cues from eye movement in a simulated scenario of an attack by a mechanical replica of the human eye. The authors further investigated in this line and published more iris liveness detection methods, based on eye movements and gaze estimation, in later years~\cite{174_15, 175_14, 178_15}. \par
Liveness-detection methods are considered temporal-based and can be easily fooled by a replay video of the face or iris or by adding liveness properties to a still face image by cutting the eye/mouth parts.

\subsection{Recapturing-detection / contextual information}
Another approach that belongs to the temporal-based category are methods that simply detect if the biometric data is a replay of a recorded sample, referred to as recapturing-detection. These methods analyze the scene and environment to detect abnormal movements in the video due to a hand holding the presenting 2D mobile device or paper~\cite{523_16_f, 507_10_f}. Such movements are different than movement of a real face with a static background. Sometimes these techniques are referred to as motion-analysis depending on motion cues from planar objects. Some use optical flow to capture and track the relative movements between different facial parts to determine spoofing~\cite{508_09_f, 509_09_f}, while other researches use motion magnification~\cite{511_16_f, 513_13_f}, Haralick features~\cite{510_16_f} or visual rhythm analysis~\cite{535_12_f}. Other types depend on contextual analysis of the scene~\cite{531_11_f,530_13_f,521_12_f,varF_13_f} or 3d reconstruction~\cite{thrD_13_f}.

The recapture detection methods fail if the attack is done using an artifact like a mask, because no replay is present to detect. So a better and more generic approach is using texture-based methods which utilize features that can tell the difference between real live biometric data and a replay or an artifact being presented to the sensor.

\subsection{Texture-based (Hand-crafted features)}
The category of texture-based methods assumes that the texture, color and reflectance characteristics captured by genuine face images are different from those resulting from presentation attacks. For example, blur and other effects caused by printers and display devices lead to detectable color artifacts and texture patterns, which can then be explored to detect presentation attacks. 

Researchers use handcrafted image feature extraction methods to extract image features from face or iris images. They then use a classification method such as support vector machines (SVM) to classify images into two classes of bona-fide or presentation attack based on the extracted image features. Such techniques are faster than temporal or motion-based analysis, as they can operate with only one image sample, or a selection of images, instead of having to analyze a complete video sequence.

\subsubsection{Texture-reflectance}
These techniques analyze appearance properties of face or iris, such as the face reflectance or texture. The reflectance and texture of a real face or iris are different than those acquired from a spoofing material; either printed paper, screen of a mobile device, mask~\cite{538_13_f, 537_13_f, 536_13_f} or glass eye.

\subsubsection{Image Quality analysis}
Techniques that depend on analysis of image quality try to detect the presence of image distortion usually found in the spoofed face or iris image. 

\paragraph{(1) Deformation for Face}
For example, in print attacks, the face might be skewed with deformed shape if an imposter bends it while holding.

\paragraph{(2) Frequency} Based on the assumption that the reproduction of previously captured images or videos affects the frequency information of the image when displayed in front of the sensor. The attack images or videos have more blur and less sharpness than the genuine ones, and so their high frequency components are affected. Hence frequency domain analysis can be used as a face PAD method as in~\cite{518_04_f, 507_10_f, b_12_f} or frequency entropy analysis by Lee~\etal~\cite{534_13_f}. For iris, Czajka~\cite{109_13} analyzed printing regularities left in printed irises and explored some peaks in the frequency spectrum. Frequency-based approaches might fail in case of high-quality spoof images or videos being presented to the camera.

\paragraph{(3) Quality metrics} Some other methods explore the potential of quality assessment to identify real and fake face or iris samples acquired from a high quality printed image. This is assuming that many image quality metrics are distorted by the display device or paper. Some researches used individual quality-based features, while others used a combination of quality metrics to better detect spoofing and differentiate between bona-fide and attack iris~\cite{104_12, 108_09, 119_07, 155_14, 164_14} or face~\cite{514_16_f, 515_15_f, 516_14_f} or both~\cite{500_18_fi, 505_17_fi}. For example, Gabally~\etal~\cite{104_12} investigated 22 iris-specific quality features and used feature selection to choose the best combination of features that discriminate between live and fake iris images. Later, in a follow-up work~\cite{565_14_fi}, they assessed 25 general image-quality metrics and not only applied on iris PAD but also used the method for face and fingerprint PAD. 
Other different approach by Garicia~\etal~\cite{524_15_f} was to analyze moire patterns which may be produced during image acquisition then display on mobile devices.

\paragraph{(4) Color analysis} Another type of algorithms utilize the fact that the distribution of color in images taken from bona-fide face or iris, is different than the distribution found in images taken from printed papers or display device. Methods for face PAD adopt solutions in a different input domain than RGB space in order to improve the robustness to illumination variation, like HSV and YCbCr color space~\cite{515_15_f, 550_15_f, 545_16_f, 546_17_f, compet_f3} or Fourier spectrum~\cite{518_04_f}. For iris PAD, color adaptive quantized patterns were used in~\cite{187_16} instead of using a gray-textured image. Z. Boulkenafeta~\etal~studied the generalization of color-texture analysis methods in~\cite{555_18_f}. 

\subsubsection{Local-descriptors}
A different group of methods use hand-crafted features which extract local-texture for PAD~\cite{533_11_f,543_13_f}. A large variety of local image descriptors have been compared with respect to their ability to identify spoofed iris, fingerprint, and face data~\cite{544_15_fi} 
and some highly successful texture descriptors have been extensively used for this purpose. For example, in face PAD methods, descriptors such as local binary pattern (LBP)~\cite{552_13_f,540_13_f, 532_11_f,547_12_f,548_13_f,549_12_f,556_15_f}, SIFT~\cite{553_16_f}, SURF~\cite{546_17_f}, HoG~\cite{548_13_f,549_12_f,541_11_f,530_13_f,310_15_f}, DoG~\cite{554_11_f,507_10_f} and GLCM~\cite{541_11_f} have been studied showing good results in discriminating real from attack face images. For PAD in iris recognition, boosted LBP was first used by He~\etal~\cite{114_09}, weighted LBP by Zhang~\etal~\cite{124_10} for contact lens detection and later, binarized statistical image features (BSIF)~\cite{161_15,171_15}.
Local phase quantization (LPQ) was investigated for biometric PAD in general, including both face and iris, by Gragnaniello~\etal~\cite{544_15_fi}, and census transform (CT) was used for mobile PAD in~\cite{558_14_fi}. Such local texture descriptors were then used with a traditional classifier like SVM, LDA, neural networks and Random Forest. 

\subsubsection{Drawback of hand-crafted texture-based methods}
Results of above methods show that the manually engineered features are suitable for solving the PAD problem for face and iris recognition systems. However, their drawback is that the design and selection of handcrafted feature extractors is mainly based on expert knowledge of researchers on the problem. Consequently, these features often only reflect limited aspects of the problem and are often sensitive to varying acquisition conditions, such as camera devices, lighting conditions and presentation attack instruments (PAIs). This causes their detection accuracy to vary significantly among different databases, indicating that the handcrafted features have poor generalizability and so do not completely solve the PAD problem. The available cross-database tests in the literature suggest that the performance of hand-engineered texture-based techniques can degrade dramatically when operating in unknown conditions. Leading to the need of automatically extracting vision meaningful features directly from the data using deep representations to assist in the task of presentation attack detection.

\subsection{Fusion}
The presentation attack detection accuracy can be enhanced by using feature-level or score-level fusion methods to obtain a more general countermeasure effective against a various types of presentation attack. For face PAD, Tronci~\etal~\cite{542_11_f} propose a linear fusion at a frame and video level combination between static and video analysis. In~\cite{541_11_f}, Schwartz~\etal~introduce feature level fusion by using Partial Least Squares (PLS) regression based on a set of low-level feature descriptors. Some other works~\cite{540_13_f, 543_13_f} obtain an effective fusion scheme by measuring the level of independence of two anti-counterfeiting systems. While Feng~\etal~in~\cite{311_16_f} integrated both quality measures and motion cues then used neural networks for classification. 

In the case of iris anti-spoofing, Z. Akhtar~\etal~\cite{558_14_fi} proposed to use decision-level fusion on several local descriptors in addition to global features. More recently Nguyen~\etal~\cite{305_18_i,306_18_i} explored both feature-level and score-level fusion of hand-crafted features with CNN extracted features, and obtained least error using the proposed score-level fusion methodology.

\subsection{Smartphones}
In the last few years, research based on smartphone and mobile device PAD is gaining popularity. Many different algorithms have been developed and datasets were released to test their performance. These databases became publicly available and we will list in this section some of these datasets together with the algorithm proposed by the authors. 

One of the first databases that was dedicated for face PAD on smartphones is MSU-MFSD~\cite{515_15_f}, the authors Wen~\etal~used a PAD method depending on image distortion analysis. Later, the same group released MSU-USSA~\cite{553_16_f} and used a combination of texture analysis (LBP and color) and image analysis techniques. 
In 2016, the IDIAP Research Institute released the Replay-Mobile~\cite{564_16_f} database and again a combinations of the Image Quality Analysis and Texture Analysis (Gabor-jets) were used as the proposed PAD.

In the field of Iris spoofing, MobBioFake dataset~\cite{155_14,166_14} was presented in 2014 that was the first to use visible spectrum instead of near-infrared for iris spoofing where the acquisition sensor is a mobile device. Sequeira~\etal~\cite{155_14} used texture-based iris PAD algorithm, they combined several image quality features at the feature level and used SVM for classification. Gragnaniello~\etal~\cite{154_15} proposed a solution based on local-descriptors but with very low complexity and required low CPU power suitable for mobile applications. They evaluated on MobBioFake and MICHE~\cite{559_14_fi} datasets and achieved good performance.

\subsection{Learned-features}
In recent years, deep learning has evolved and the use of deep neural networks or convolutional neural networks (CNN) has proved to be effective in many computer vision tasks especially with the availability of new advanced hardware and large data. CNNs have been successfully used for vision problems like image classification and object detection. This has encouraged many researchers to incorporate deep learning in the PAD problem. Instead of using hand-crafted features, rely on a deep network or CNN to learn features that discriminate between bona-fide and attack face~\cite{311_16_f, 313_14_f, 315_16_f, 318_16_f, 319_18_f, 314_17_f, 316_17_f, 317_18_f} or iris~\cite{305_18_i, 306_18_i, 307_17_i, 309_15_i} or both~\cite{308_15_fi, 312_18_fi}. Some proposed to use hybrid features that combine information from both handcrafted and deeply-learnt features.

Several works use a pretrained CNN as a feature extractor, then use a normal classifier as SVM for classification, like in~\cite{315_16_f}. Li~\etal~\cite{315_16_f} finetune a pretrained VGG CNN model on ImageNet~\cite{p1_15}, use it to extract features, reduce dimensionality by principal component analysis (PCA) then classify with SVM. Combining deep features with motion cues was proposed in several face video spoof detection. In~\cite{318_16_f}, Patel~\etal~finetuned a pretrained VGG for texture features extraction in addition to using frame difference as motion cue. Similarly, Nguyen~\etal~\cite{304_18_f} used VGG for deep feature extraction then fused it with multi-level LBP features, and used SVM for final classification. Feng~\etal~\cite{311_16_f} combine image quality information together with motion information from optical flow as input to a neural network for classification of bona-fide or attack face.  Xu~\etal propose an LSTM-CNN architecture to utilize temporal information for binary classification in~\cite{310_15_f}.

Atoum~\etal~\cite{301_17_f} propose a two-steam CNN-based face PAD method using texture and depth, they utilized both full face and patches from the face with different color spaces instead of RGB. They evaluated on three benchmark databases but did not perform cross-dataset evaluation. Later, Liu~\etal~\cite{300_18_f} extended and refined this work by learning another auxiliary information, rPPG signal, from the face video and evaluated their depth-supervised learning approach on more recent datasets OULU-NPU~\cite{oulu} and SiW. Wang~\etal~\cite{302_18_f} did not consider depth as an auxiliary supervision like in~\cite{300_18_f}, however, they used multiple RGB frames to estimate face depth information, and used two modules to extract short and long-term motion. Four benchmark datasets were used to evaluate their approach.

Use of CNN by itself for the sake of classifying attack vs bona-fide iris images has been investigated in recent years. Andrey K.~\etal~\cite{320_18_i} proposed to use a lightweight CNN on binarized statistical image features extracted from the iris image, they evaluated their proposed algorithm on LivDet-Warsaw 2017~\cite{182_17} and other 3 benchmark datasets. In~\cite{321_18_i} Hoffman~\etal~proposed to use 25 patches of the iris image as input to the CNN instead of the full iris image. Then a patch-level fusion method is used to fuse scores from the input patches based on the percentage of iris and pupil pixels inside each patch. They performed cross-dataset evaluation on LivDet-Iris 2015 Warsaw, CASIA-Iris-Fake, and the BERC-Iris-Fake datasets and used true detection rate (TDR) at 0.2\% false detection rate (FDR) as their evaluation metric.

%-------------------------------------------------------------------------
\section{Competitions and Databases}\label{sec:compet_db}
In this section, we summarize the face and iris PAD competitions held in addition to benchmark datasets used to evaluate performance of the face or iris presentation attack detection solutions. All these databases are publicly available upon request from the authors. Table~\ref{tabDB} shows summary of information about all six datasets used.

\subsection{Competitions}
Since 2011, several competitions were held to assess the status of presentation attack detection for either face or iris. Such competitions were very useful for collecting new benchmark datasets and creating a common framework for evaluating the different PAD methods.

For face presentation attack detection, three competitions were carried out since 2011. The first competition was held during 2011~\cite{compet_f1} on the IDIAP Print-Attack dataset~\cite{533_11_f} with six competing teams, the winning methodology was using hand-engineered texture-based approaches which proved to be very effective to detect print attacks. After two years, the second face PAD competition was carried out in 2013~\cite{compet_f2} with 8 participating teams, on the Replay-Attack dataset~\cite{547_12_f}. Most of the teams relied only on texture-based methods, while 3 teams used hybrid approaches combining both texture and motion based countermeasures. Two of these hybrid approaches reached 0\% HTER on the test set of the Replay-Attack database.

Later in 2017, a competition was held on generalized software-based face presentation attack detection in mobile scenarios~\cite{compet_f3}. Thirteen teams participated and four protocols were used for evaluation. The winning team that showed best generalization to unseen cameras and attack types, used a combination of color, texture and motion-based features.

More recently, after the release of the large-scale multi-modal face anti-spoofing dataset, CASIA-SURF~\cite{303_19_f}; the Chalearn LAP Multi-Modal Face anti-spoofing Attack Detection Challenge~\cite{chalearn_19_f} was held in 2019.
Thirteen teams were qualified for the final round with all submitted face PAD solutions relying on CNN-based feature extractors. The top 3 teams utilized ensembles of feature extractors and classifiers. For future work, the challenge paper suggested to take full advantage of the multi-modal dataset, by defining a set of cross-modal testing protocols, as well as introducing 3D mask attacks in addition to the 2D attacks. These recommendations were later fulfilled in the following year, at the Chalearn Multi-modal Cross-ethnicity Face anti-spoofing Recognition Challenge, using the CASIA-SURF CeFA~\cite{db_19_f} dataset.

For iris presentation attack detection, the first competition was LivDet-Iris 2013~\cite{159_14} then two more sequels for this competition were held in 2015~\cite{181_15} and 2017~\cite{182_17}. Images in LivDet-Iris included iris printouts and textured contact lenses which were more difficult to detect than printouts. For LivDet-Iris 2017, the test images were split into known and unknown parts, where the known part contained images that were taken under similar conditions with the same sensor as training images. On the other hand, the unknown partition consisted of images taken in different environment with different properties. Results on known partition were much better than those on the unknown part. In 2014, another competition was held in 2014 specifically for mobile authentication; Mobile Iris Liveness Detection Competition (MobILive 2014)~\cite{156_14}. A mobile device was used to capture images of iris printouts in visible light; MobBioFake dataset, not infra-red light as datasets used in LivDet-Iris competitions. Six teams participated and the winning team achieved perfect detection of presentation attack. For a more thorough assessment of these iris competitions, refer to Czajka and Bowyer~\cite{183_18}.

\begin{table}[!t]
\processtable{Properties of used face and iris databases.\label{tabDB}}
{\begin{tabular*}{20pc}{@{\extracolsep{\fill}}llll@{}}
\toprule

\multirow{2}{*}{Database} & Number of samples & Resolution &  FPS / Video  \\
& (Attack + Bona-fide) & ( W * H )  & duration (face videos) \\
\midrule
%\hline %\hline
Replay-Attack (Face) & $1000 + 200$ (videos) & $320 \times 240$ & 25 / 9-15 seconds \\
MSU-MFSD (Face) & $210 + 70$   (videos) & $720 \times 480$ & 30 / 9-15 seconds \\
Replay-Mobile (Face) & 640 + 390  (videos) & $720 \times 1280$ & 30 / 10-15 seconds \\
Warsaw 2013 (Iris) & $815 + 825$  (images) & $640 \times 480$ & - \\
ATVS-FIr (Iris)&  $800 + 800$  (images) & $640 \times 480$ & - \\
MobBioFake (Iris) & $800 + 800$  (images) &  $250 \times 200$ & - \\

\botrule
\end{tabular*}}{}
\end{table}

\begin{table*}[t!]
\fwprocesstable{Face databases details. \label{tab:DBface}}
{\begin{tabular*}{\textwidth}{C{1.6cm}L{2.8cm}L{3.5cm}L{2.5cm}L{1.2cm}L{1.2cm}L{1.2cm}L{1.8cm}}

		\toprule
Dataset	 & Sensor used for authentication	& PA artifact	& PAI (Presentation Attack Instrument)		& Lighting conditions (bona-fide / attack)	& Subjects (train / dev / test)	& Videos per subject (bona-fide + attack) & Videos per subset	(bona-fide + attack)\\		

\midrule

Replay-Attack (2012)
& \begin{tabular}[t]{@{}p{2.8cm}@{}} (1) \textbf{W} in MacBook laptop ($320\times240$) 
  \end{tabular} 
& \begin{tabular}[t]{@{}l@{}} 
    1) \textbf{PH} $\star$ \\ 
    2) 720p \textbf{high-def V} $\star$ \\
    $\star$: Canon PowerShot SX150 IS\end{tabular} 
& \begin{tabular}[t]{@{}l@{}} 
    1) \textbf{PR} (A4) \\ 
    2) \textbf{VR} on \textbf{M} (iPhone)\\ 
    3) \textbf{VR} on \textbf{T} (iPad) \end{tabular} 
& 2 / 2 
& 50 (15/15/20)

& 4 + 20
& \begin{tabular}[t]{@{}l@{}l@{}} 
    Train & : 60+300\\ 
    Dev & : 60+300 \\ Test & : 80+400
   \end{tabular}  \\

\midrule

CASIA-FASD (2012)
& \begin{tabular}[t]{@{}p{2.8cm}@{}} 
    1) Low \textbf{Q}: longtime-used \textbf{USB \textbf{C}} ($680\times480$) \\ 
    2) Normal \textbf{Q}: new \textbf{USB \textbf{C}} ($680\times480$) \\ 
    3) High \textbf{Q}: Sony NEX-5 ($1920\times1080$)     \end{tabular}
& \begin{tabular}[t]{@{}l@{}} 
    1) \textbf{PH} $\star$  \\ 
    2) 720p \textbf{high-def V} $\star$ \\ 
     $\star$: Sony NEX-5
   \end{tabular}
& \begin{tabular}[t]{@{}p{2.5cm}@{}}
 1) \textbf{PR} (copper A4) - Warped \\ 
 2) \textbf{PR} (copper A4) - Cut  \\ 
 3) \textbf{VR} on \textbf{T} (iPad)
 \end{tabular}
& 1 / 1
& 50 (20/-/30) 
& 3 + 9
& \begin{tabular}[t]{@{}l@{}l@{}} 
    Train & : 60+180 \\ 
    Test &: 90+270   
 \end{tabular}
\\		
						
\midrule																		
	
MSU-MFSD (2014)
& \begin{tabular}[t]{@{}p{2.8cm}@{}} 
    1) \textbf{W} in MacBook Air ($640\times480$)  \\
    2) \textbf{FC} of Google Nexus 5 \textbf{M} ($720\times480$) \end{tabular}
& \begin{tabular}[t]{@{}p{3.5cm}@{}} 
    1) \textbf{PH} (Canon PowerShot 550D SLR) \\
    2) 1080p \textbf{high-def V} (Canon PowerShot 550D SLR) \\ 
    3) 1080p \textbf{M V} (\textbf{BC} of iPhone S5)
 \end{tabular}
& \begin{tabular}[t]{@{}p{2.5cm}@{}}  
  1) \textbf{PR} (A3) \\ 
  2) \textbf{high-def VR} on \textbf{T} (iPad) \\ 
  3) \textbf{M VR} on \textbf{M} (iPhone) \end{tabular}
& 1 / 1
& 35 (15/-/20)
& 2 + 6
& \begin{tabular}[t]{@{}l@{}l@{}} 
    Train & : 30+90 \\ 
    Test & : 40+120 
 \end{tabular} \\

\midrule																			

Replay-Mobile (2016)
& \begin{tabular}[t]{@{}p{2.8cm}@{}}
    1) \textbf{FC} of iPad Mini2 \textbf{T} ($720\times1280$)\\
    2) \textbf{FC} of  LG-G4 \textbf{M} ($720\times1280$)
  \end{tabular}
& \begin{tabular}[t]{@{}p{3.5cm}@{}}
    1) \textbf{PH} (Nikon coolix P520) \\
    2) 1080p \textbf{M V} (\textbf{BC} of LG-G4 \textbf{M})
  \end{tabular}
& \begin{tabular}[t]{@{}p{2.5cm}@{}}
    1) \textbf{PR} (A4) \\
    2) \textbf{VR} on matte-screen (Philips 227ELH screen) 
   \end{tabular}
&  5 / 2
& 40 (12/16/12)
&  10 + 16	
& \begin{tabular}[t]{@{}l@{}l@{}} 
    	Train & : 120+192 \\
    	Dev & : 160+256 \\
    	Test & : 110+192
 \end{tabular} \\

\midrule																		

OULU-NPU (2017)
& \textbf{FC} of 6 \textbf{M} ($1920\times1080$)

& \begin{tabular}[t]{@{}p{3.5cm}@{}}
    1) \textbf{PH} $\star$ \\
    2) \textbf{high-res M V} $\star$ \\
    $\star$: \textbf{BC} of Samsung Galaxy S6 Edge \textbf{M}
  \end{tabular}
& \begin{tabular}[t]{@{}p{2.5cm}@{}}
    1-2) \textbf{PR} (glossy A3) - 2 Printers \\
    3) High-res \textbf{VR} on 19 inch Dell display  \\ 
    4) High-res \textbf{VR} on 13 inch Macbook
   \end{tabular}
& 3 / 3
&  55 (20/15/20) 
& 36+72 
& \begin{tabular}[t]{@{}p{1.8cm}@{}}
    4 Protocols \\
    Total: 1980+3960
 \end{tabular} \\
\midrule																

SiW (2018)
& \begin{tabular}[t]{@{}p{2.8cm}@{}}  
        1) High \textbf{Q C}: Canon EOS T6 ($1920\times1080$)\\
        2) High \textbf{Q C}: Logitech C920 \textbf{W} ($1920\times1080$)\\

    \end{tabular}
& \begin{tabular}[t]{@{}p{3.5cm}@{}}
    1) \textbf{PH} (5, 184 x 3, 456) \\
    2) Use same live videos \\
  \end{tabular}
& \begin{tabular}[t]{@{}p{2.5cm}@{}}
    1) \textbf{PR} \\
    2) \textbf{PR} of frontal-view frame from a live \textbf{V} \\
    3-6) \textbf{VR} on 4 screens: Samsung S8, iPhone7, iPad Pro, PC
   \end{tabular}
& 1 / 1
& 165 (90/-/75)
& 8 + 20
& \begin{tabular}[t]{@{}p{1.8cm}@{}}
    3 Protocols \\
    Total: 1320+3300
 \end{tabular} \\

\midrule																		

CASIA-SURF (2018)
& \begin{tabular}[t]{@{}p{2.8cm}@{}}
    1) RealSense SR300 \\
    - RGB: ($1280\times720$) \\
    - Depth/IR: ($640\times480$)
  \end{tabular}
& \begin{tabular}[t]{@{}p{3.5cm}@{}}
    1) \textbf{PH}
  \end{tabular}
& \begin{tabular}[t]{@{}p{2.5cm}@{}}
    1) \textbf{PR} (A4)
   \end{tabular}
& 1 / 1
& \begin{tabular}[t]{@{}l@{}} 
     1000 \\  (300/100/600)
   \end{tabular}
& 1 + 6
& \begin{tabular}[t]{@{}p{1.7cm}@{}} 
    Train: 900+5400 \\
    Dev: 300+1800 \\
    Test: 1800+10800 
  \end{tabular}\\

\midrule		

CASIA-SURF CeFA (2019)
& \begin{tabular}[t]{@{}p{2.8cm}@{}}
    1) Intel Realsense \\
    - RGB/Depth/IR  \\
    - all ($1280\times720$)
  \end{tabular}
& \begin{tabular}[t]{@{}p{3.5cm}@{}}
    1) \textbf{PH} \\
    2) Same live videos \\
    3) 3D print face mask \\
    4) 3D silica gel face mask
  \end{tabular}
& \begin{tabular}[t]{@{}p{2.5cm}@{}}
    1) \textbf{PR} \\
    2) \textbf{VR} 
   \end{tabular}
& \begin{tabular}[t]{@{}l@{}}

    1 / 2 (2D) \\
    \\
    0 / 6 \\
    0 / 4
   \end{tabular} 

& \begin{tabular}[t]{@{}p{1.2cm}@{}}
    1500 (600/300/600)\\
    99 (0/0/99) \\
    8 (0/0/8)
   \end{tabular}
& \begin{tabular}[t]{@{}l@{}}

    1 + 3 \\
    \\
    0 + 18 \\
    0 + 8
 \end{tabular}
& \begin{tabular}[t]{@{}p{1.8cm}@{}}
    4 Protocols: \\
    4500+13500\\
    0+5346 \\
    0+196
 \end{tabular} \\

\botrule

\end{tabular*}}{ \textbf{C}: Camera, \textbf{BC}: Back-Camera, \textbf{FC}: Front-Camera, \textbf{W}: built-in webcam, \textbf{T}: Tablet, \textbf{M}: Mobile Smartphone, \textbf{Q}: Quality, \textbf{PH}: High-res photo, \textbf{V}: Video, \textbf{PR}: Hard-copy print of high-res photo, \textbf{VR}: Video replay}

\end{table*}

%-------------------------------------------------------------------------
\subsection{Face spoofing Databases}
In Table~\ref{tab:DBface} we add details on the presentation media and presentation attack instruments (PAI) of the used face video datasets.

\subsubsection{Replay-Attack}

\begin{figure}[b!]
    \centering
        \includegraphics[width=.3\linewidth]{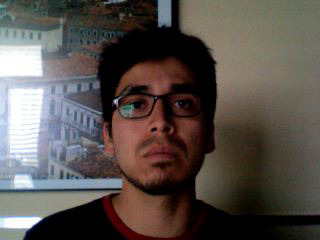}  
        \includegraphics[width=.3\linewidth]{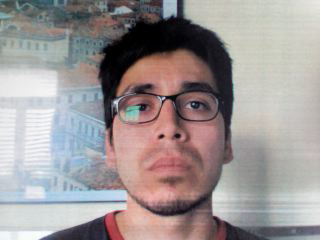}
        \includegraphics[width=.3\linewidth]{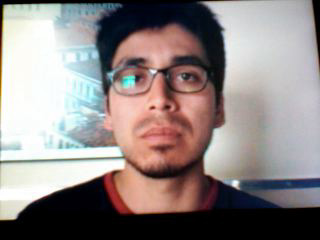}
        \includegraphics[width=.3\linewidth]{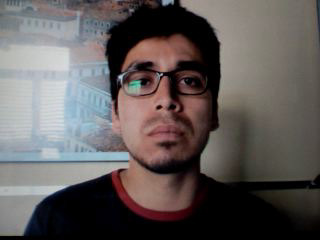}
        \includegraphics[width=.3\linewidth]{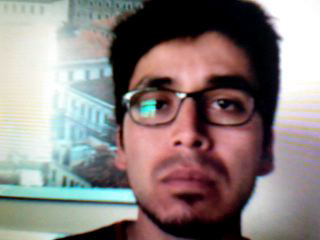}
        \includegraphics[width=.3\linewidth]{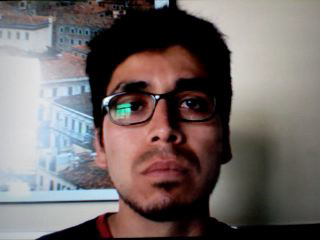}

    \caption{Sample images from the Replay-Attack datasets (adverse lighting, hand-support).
    \figfooter{*}{Left to right: top: (bona-fide, print highdef photo, mobile photo), bottom: (highdef photo, mobile video, highdef video)}
     \figfooter{*}{client 007}
}
    \label{fig:repAtt}
\end{figure}

One of the earliest datasets presented in literature in 2012 for the problem of face spoofing is the Replay-Attack dataset~\cite{547_12_f}, as a sequel to its Print-Attack version introduced in 2011~\cite{533_11_f}. The database contains a total of 1200 short videos between 9 and 15 seconds. The videos are taken at resolution $320 \times 240$ from 50 different subjects. Each subject has 4 bona-fide videos and 20 spoofed ones, so the full dataset has 200 bona-fide videos and 1000 spoofed videos. The dataset is divided into 3 subject-disjoint subsets one for training, one for development and one for testing. The training as well as the development subset each has 15 subjects, while the test subset has videos from the remaining 20 subjects. Results on this dataset is reported as the HTER (Half Total Error Rate) on the test subset when the detector threshold is set on the EER (Equal Error Rate) of the development set. For both the real and spoof videos, each subject was recorded twice with a regular webcam in two different lighting settings; controlled and adverse. 

As for the spoofed videos, first high-resolution videos were taken of the subject with a Canon PowerShot SX150 IS camera, then three techniques were used to generate 5 spoofing scenarios: (1) "hard-copy print-attack", where high-resolution digital photographs are presented to the camera. The photographs are printed with a Triumph-Adler DCC 2520 color laser printer on A4 paper. (2) "mobile-attack" using a photo once and replay of the recorded video on an iPhone 3GS screen (with resolution $480 \times 320$ pixels) once, (3) "high-definition" attack by displaying a photo or replaying the video on "high-resolution screen" using an iPad (first generation, with a screen resolution of $1024 \times 768$ pixels). Each of the 5 scenarios was recorder in two different lighting conditions as stated above, then presented to the sensor in 2 different ways. Either with a "fixed" tripod, or by an attacker holding the presenting device (printed paper or replay device) with his/her "hand", leading to a total of 20 attack videos per subject. More details of the dataset can be found at the IDIAP Research Institute website \footnote{\url{http://www.idiap.ch/dataset/replayattack}}.

\subsubsection{MSU Mobile Face Spoofing Database (MSU-MFSD)}
\begin{figure}[t!]
    \centering
%    \begin{subfigure}[t]{1\linewidth}
    \begin{subfigure}[t]{0.5\textwidth}
        \centering
        \includegraphics[width=.23\linewidth]{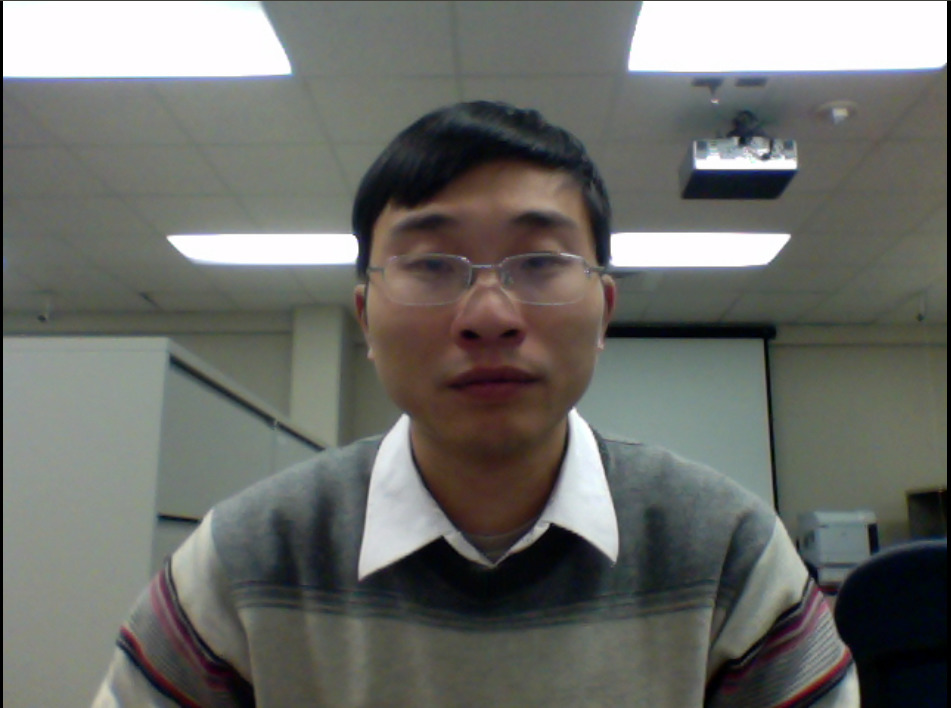}  
        \includegraphics[width=.23\linewidth]{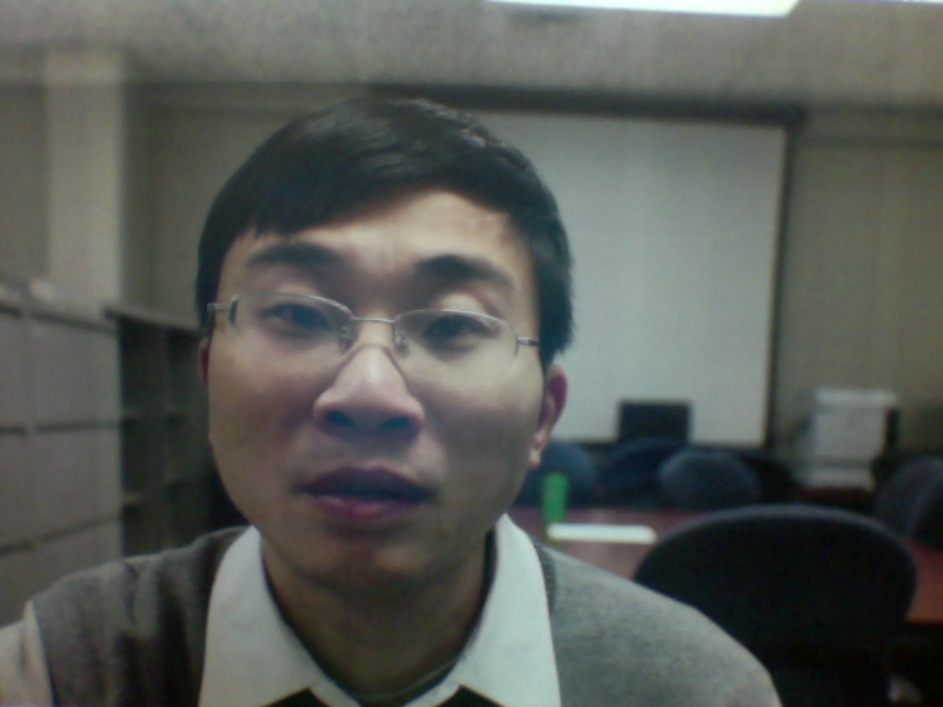}
        \includegraphics[width=.23\linewidth]{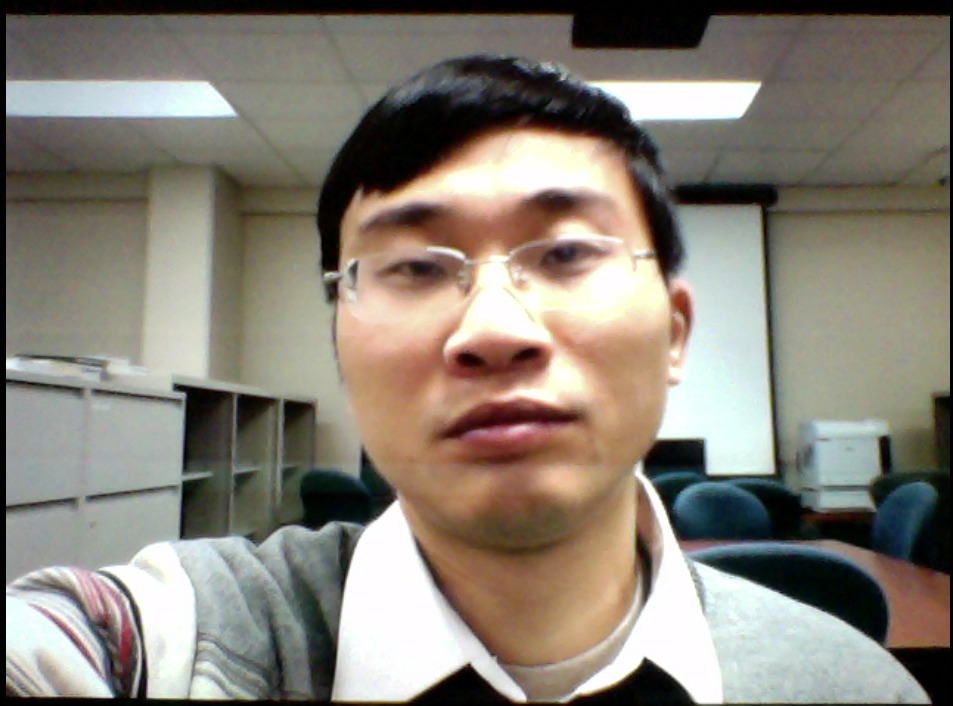}
        \includegraphics[width=.23\linewidth]{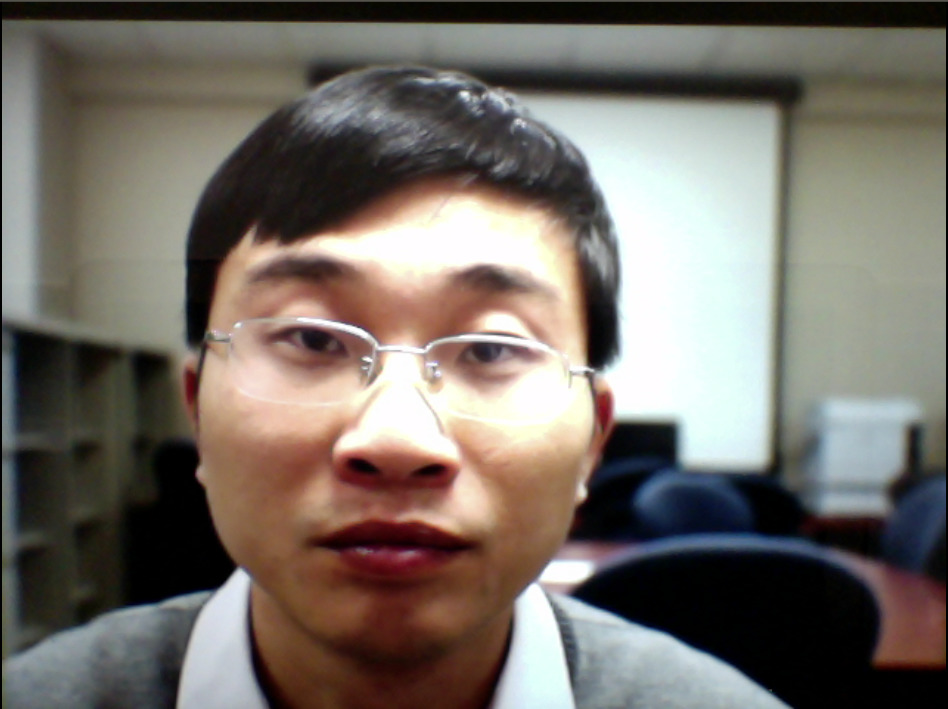}
        \caption{Laptop}
        \label{fig:msu1}
    \end{subfigure}%
    \hfill
    \begin{subfigure}[t]{0.5\textwidth}
        \centering
        \includegraphics[width=.23\linewidth]{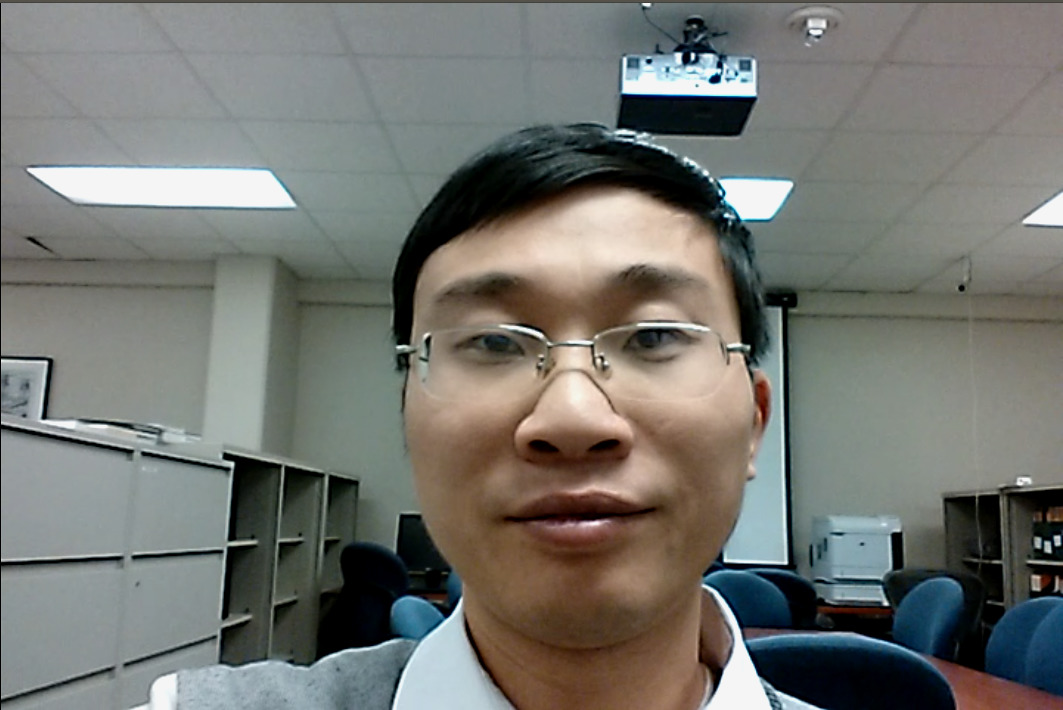}  
        \includegraphics[width=.23\linewidth]{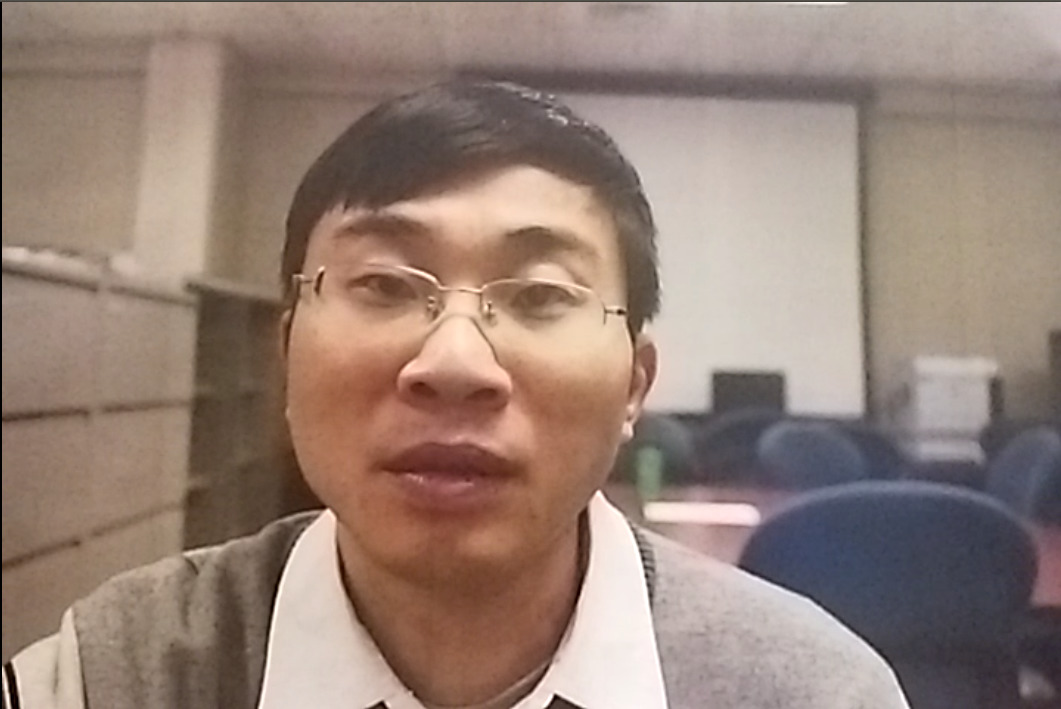}
        \includegraphics[width=.23\linewidth]{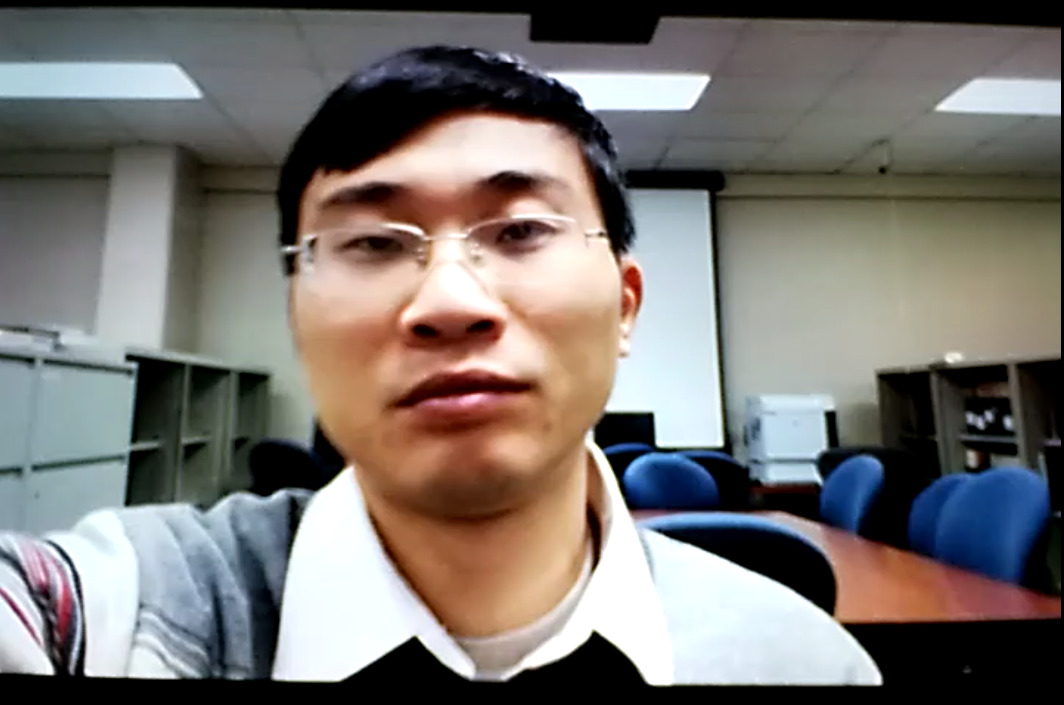}
        \includegraphics[width=.23\linewidth]{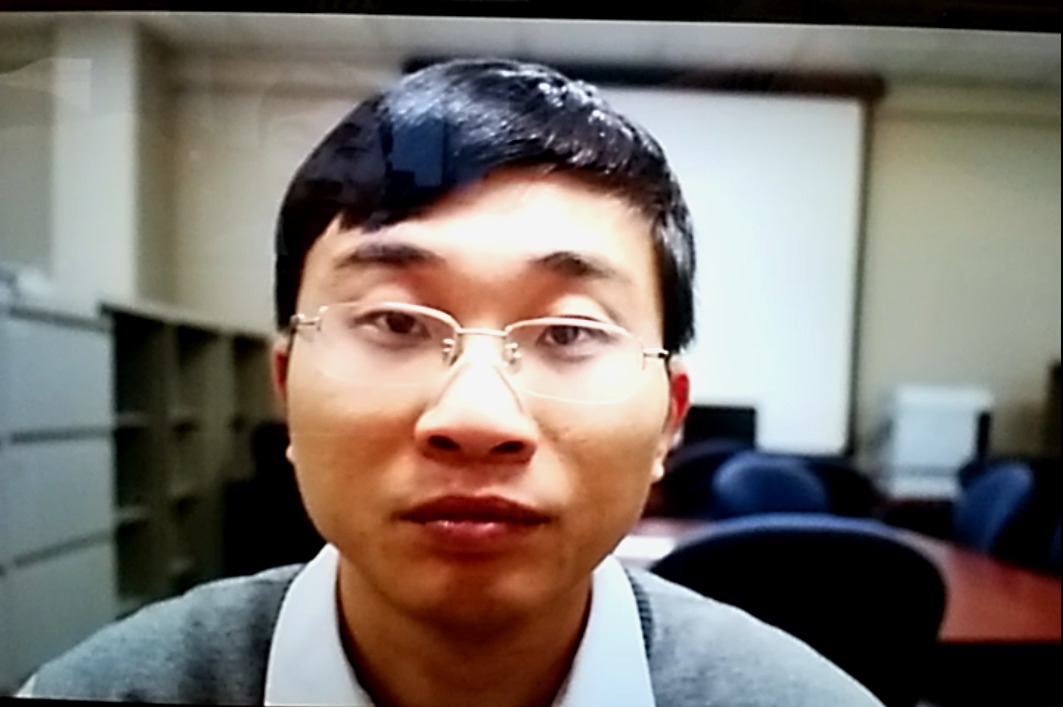}
        \caption{Android}
        \label{fig:msu2}
    \end{subfigure}

    \caption{Sample images from the MSU-MFSD datasets.
    \figfooter{*}{Left to right: bona-fide, printed photo, iphone video, ipad video}
    \figfooter{*}{client 001}}
    \label{fig:msu}
%\end{figure*}
\end{figure}

The MSU-MFSD dataset~\cite{515_15_f} was introduced in 2014 by the Michigan State University (MSU). It's main aim was to tackle the problem of face spoofing on smartphones where some mobile phone applications use face for authentication and unlocking the phone. The dataset includes real and spoofed videos from 35 subjects with a duration between 9 and 15 seconds. Unlike Replay-Attack dataset where only a webcam was used for authentication, in MSU-MFSD two devices were used, the built-in webcam of a MacBook Air 13" with resolution $640 \times 480$ and the front facing camera of the Google Nexus 5 smartphone with $720 \times 480$ resolution. So for each subject there are 2 bona-fide videos, one from each device, and six attack videos, 3 from each device. Leading to a total of 280 videos, 210 attack and 70 bona-fide videos. The attack scenarios are all presented to the authentication sensor with a fixed support unlike Replay-Attack which has videos with fixed and others using hand-held devices. Three attack scenarios were used (1) print-attack on A3 paper (2) video replay attacks on the screen of an iPad Air and (3) video replay attacks on Google Nexus 5 smartphone. The dataset is divided into two subject-disjoint subsets, one for training with 15 subjects, and the other for testing with 20 subjects. The EER on the test set is used as the evaluation metrics on the MSU-MFSD database.

\subsubsection{Replay-Mobile}

\begin{figure}[b!]
    \centering
%    \begin{subfigure}[t]{1\linewidth}
    \begin{subfigure}[t]{0.5\textwidth}
        \centering
        \includegraphics[width=.23\linewidth]{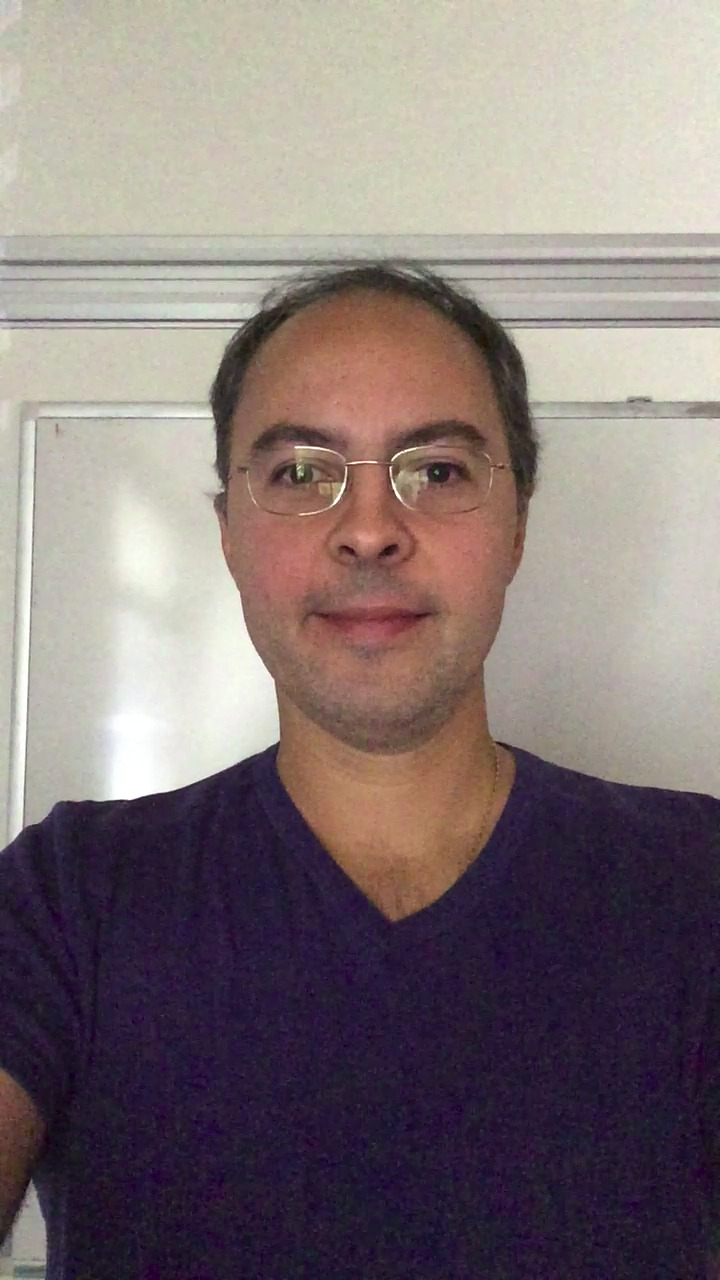}  
        \includegraphics[width=.23\linewidth]{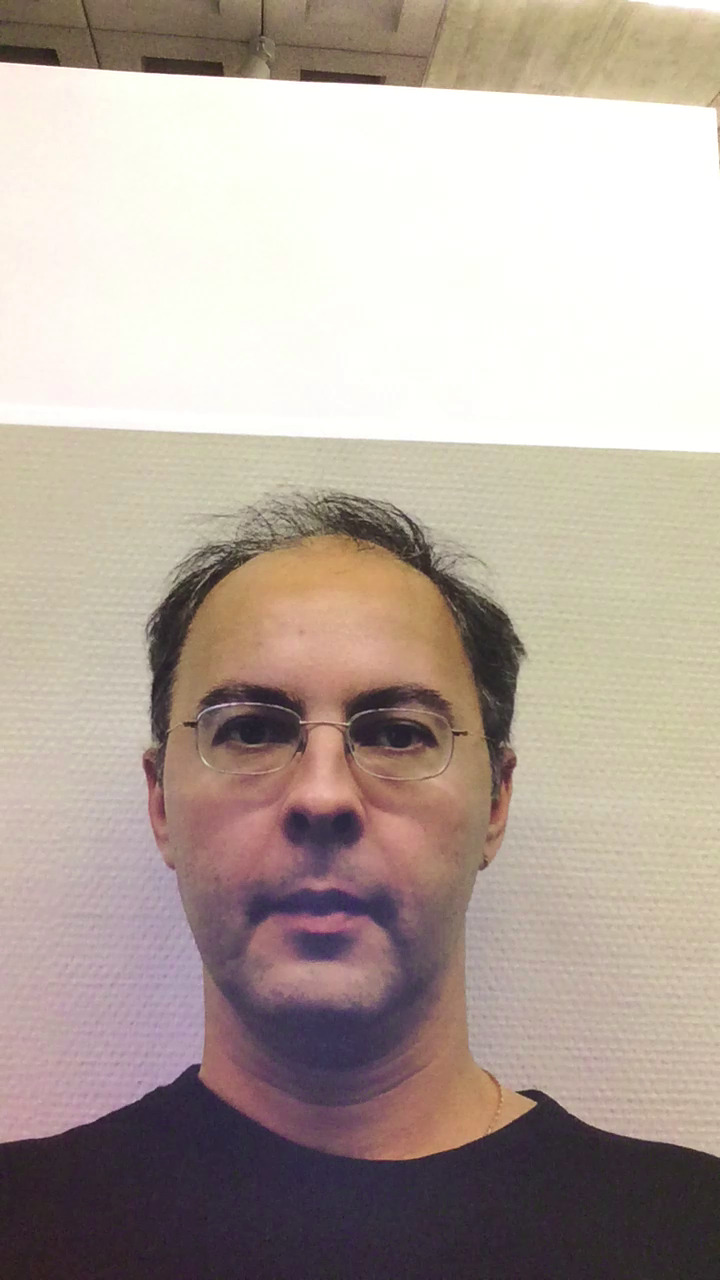}
        \includegraphics[width=.23\linewidth]{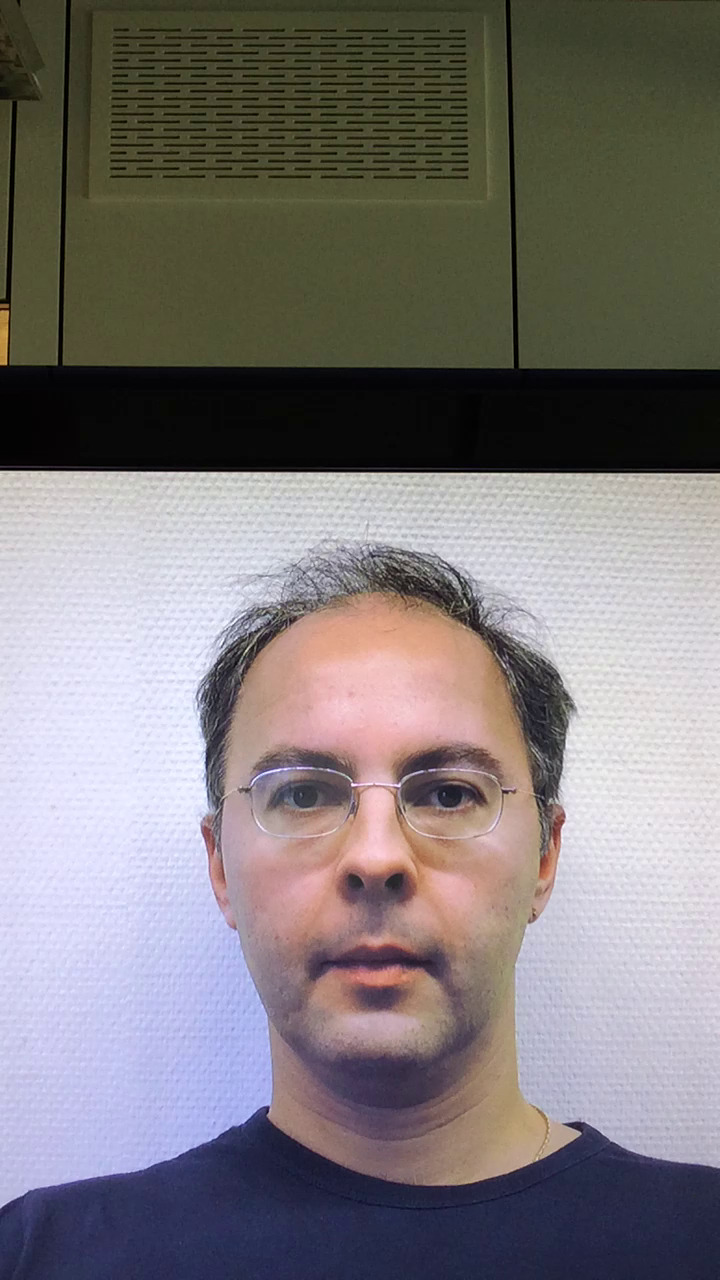}
        \includegraphics[width=.23\linewidth]{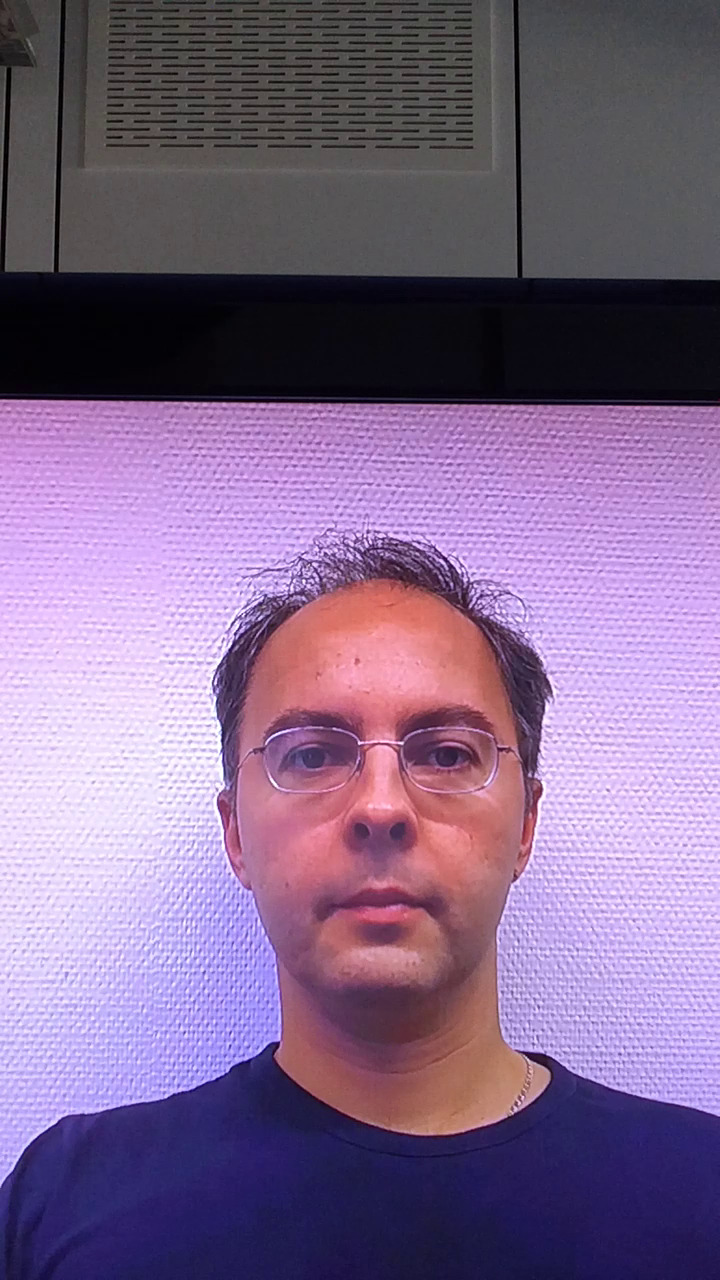}
        \caption{Captured by Tablet}
        \label{fig:repMob1}
    \end{subfigure}%
    \hfill
    \begin{subfigure}[t]{0.5\textwidth}
        \centering
        \includegraphics[width=.23\linewidth]{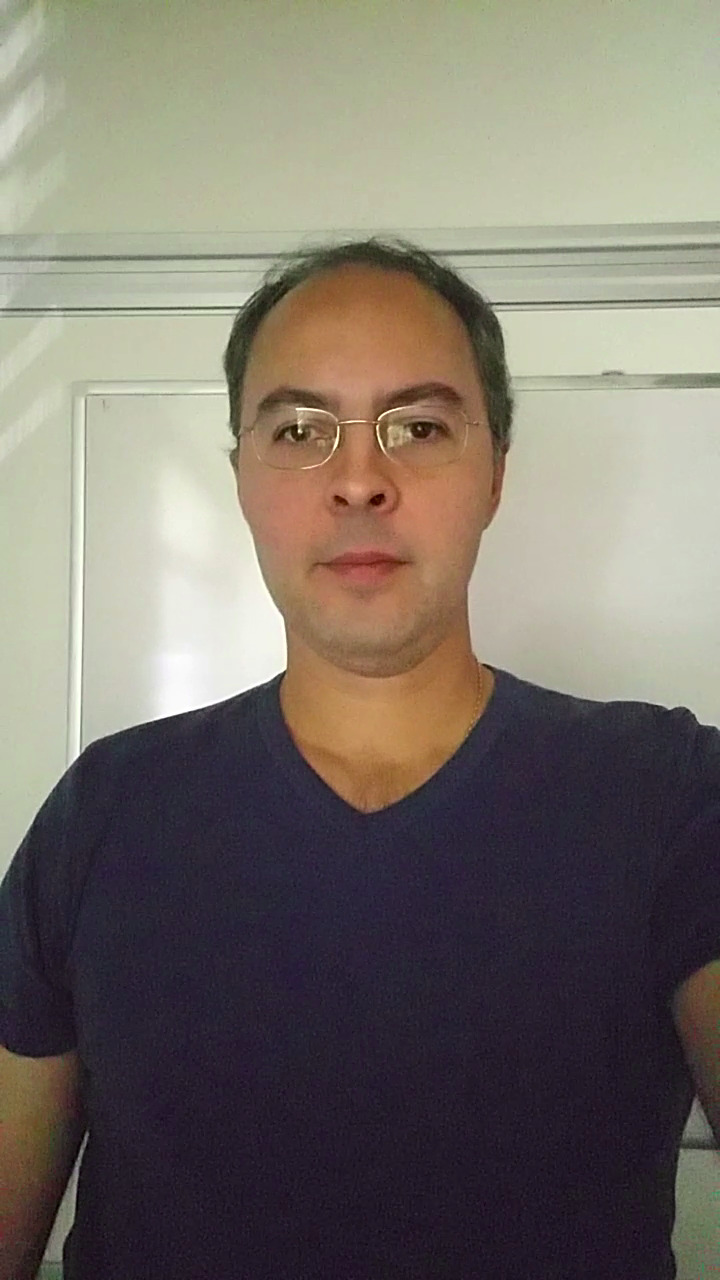}  
        \includegraphics[width=.23\linewidth]{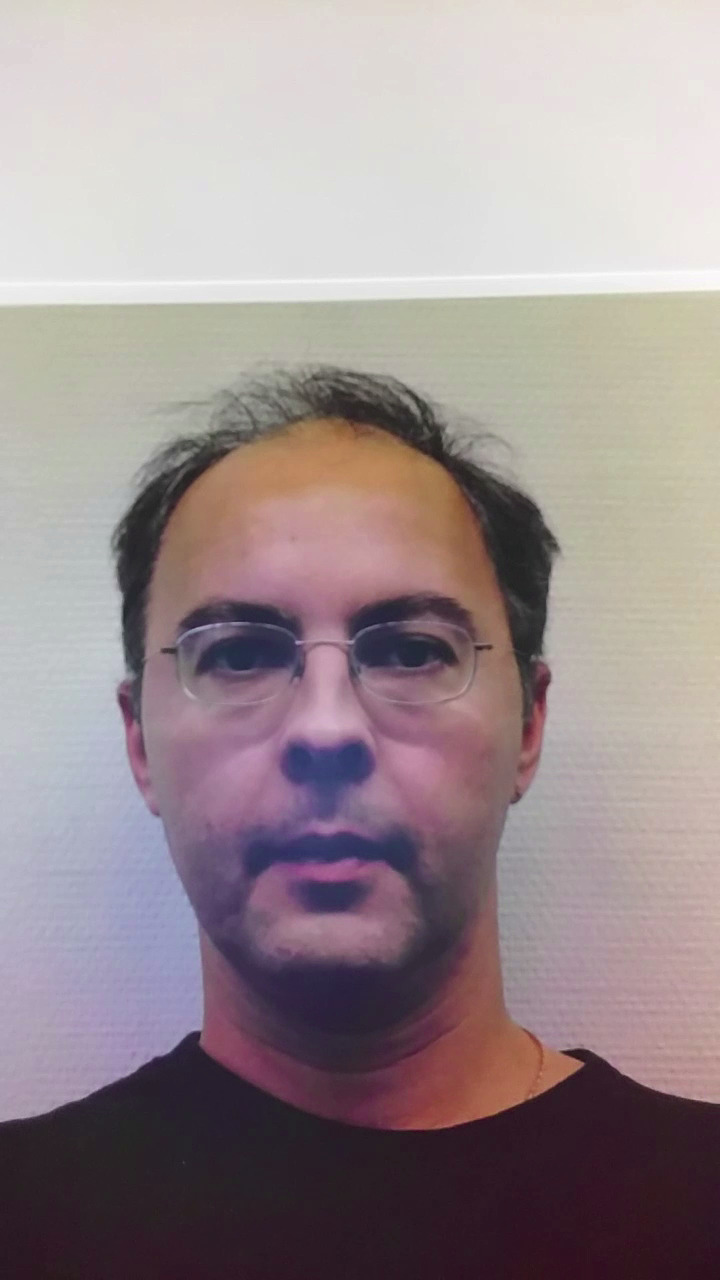}
        \includegraphics[width=.23\linewidth]{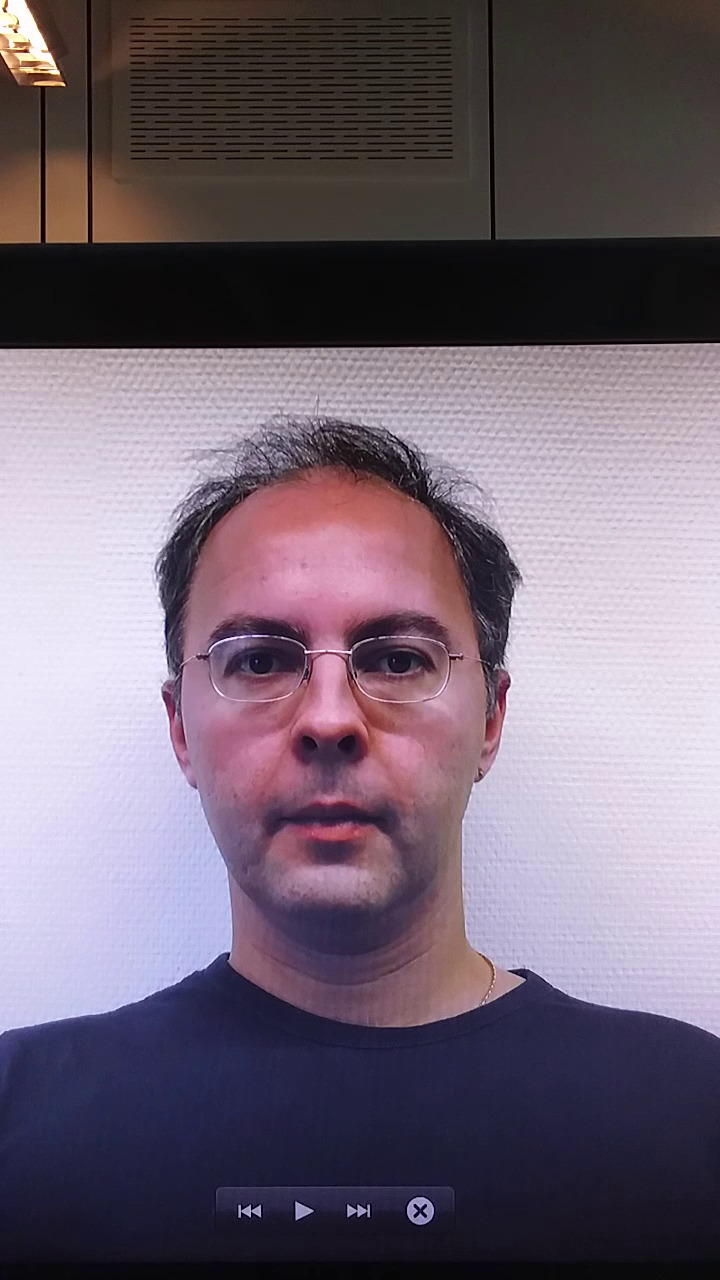}
        \includegraphics[width=.23\linewidth]{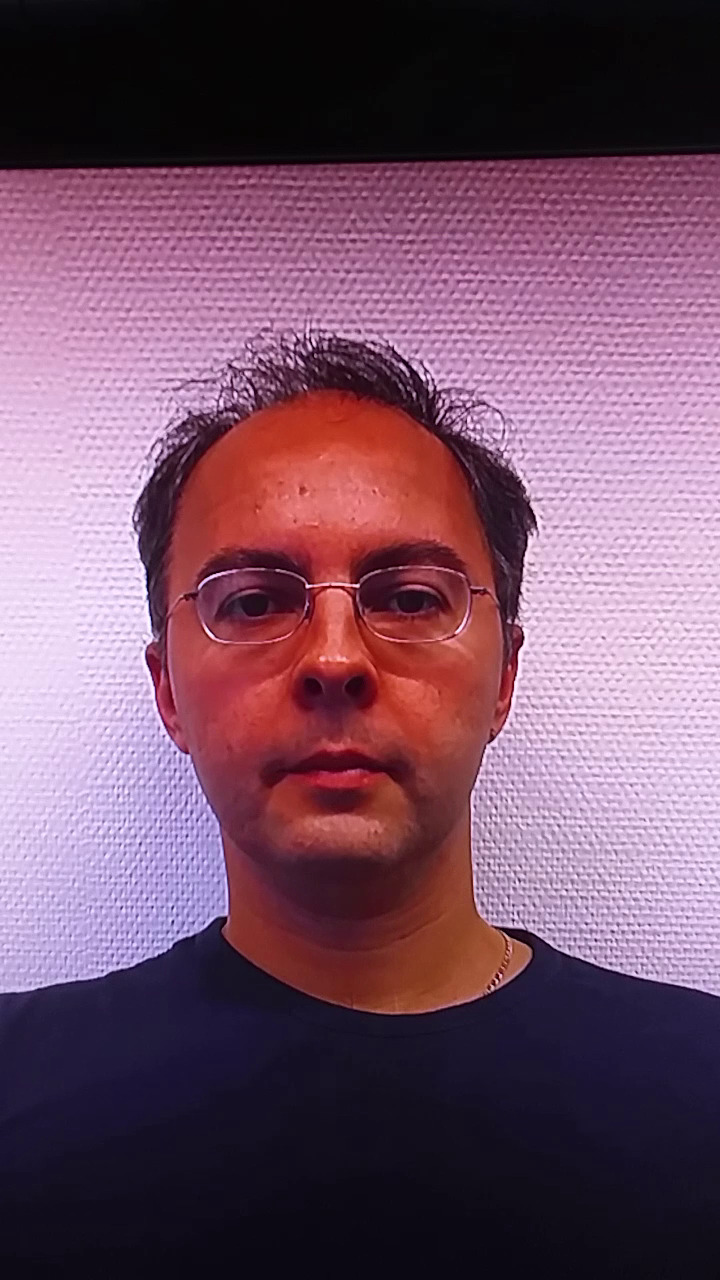}
        \caption{Captured by Mobile}
        \label{fig:repMob2}
    \end{subfigure}

    \caption{Sample images from the Replay-Mobile datasets (hand-support).
    \figfooter{*}{Left to right: bona-fide, print photo, mattescreen photo, mattescreen video)}
     \figfooter{*}{client 001}
}
    \label{fig:repMob}
\end{figure}

In 2016, the IDIAP research institute that released Replay-Attack, released a new dataset Replay-Mobile~\cite{564_16_f}. This was after the widespread of face authentication on mobile devices and the increase of the need for face-PAD evaluation sets to capture the characteristics of mobile devices. The dataset has 1200 short recordings from 40 subjects captured by two mobile devices at resolution $720 \times 1280$. Ten bona-fide accesses were recorded for each subject, in addition to 16 attack videos taken under different attack modes. Like Replay-Attack, the dataset was divided into training, development and testing subsets of 12, 16 and 12 subjects respectively, and results are reported as HTER on the test set.

Five lighting condition were used in recording real-accesses; controlled, adverse, direct, lateral and diffuse). For presentation attack, high-resolution photos and videos from each client were taken under two different lighting conditions (lighton, lightoff).

Attack was performed in two ways, (1) photo-print attacks where the printed high-resolution photo was presented to the capturing device using either fixed or hand support, (2) matte-screen attack displaying digital-photo or video with fixed supports.

%-------------------------------------------------------------------------
\subsubsection{Others} % "not used in this paper's experiments"}
Some other datasets for face PAD are also available but were not used in this paper.
\paragraph{CASIA-FASD~\cite{b_12_f}} was released in 2012, having 50 subjects. Three different imaging qualities were used (Low, Normal and High), and three types of attacks are generated from the high quality records of the genuine faces. Namely, cut-photo attack, warped-photo attack and video attack on iPad were used. The dataset is split into training set of 20 subjects, and a testing set including the other 30 subjects.

\paragraph{OULU-NPU~\cite{oulu}} A mobile face PAD dataset was released in 2017 of 55 subjects captured in three different lighting conditions and background with six different mobile devices. Both print and video-replay attacks are included using two different PAIs each. Four protocols were proposed in the paper to evaluate the generalization of PAD algorithms against different conditions.
           
\paragraph{Spoof in the Wild (SiW)~\cite{300_18_f}} Contains 8 real and 20 attack videos for 165 subjects; more subjects that any previous dataset, and released in 2018. Bona-fide videos are captured with two high-quality cameras and collected with different orientations, facial expressions and lighting conditions. Attack samples are either high-quality print attacks, or replay videos on 4 different PA devices.

\paragraph{CASIA-SURF~\cite{303_19_f}} Later in same year; 2018, a large-scale multi-modal dataset CASIA-SURF was released containing 21000 videos for 1000 subjects. Each video has three modalities, namely, RGB, Depth and IR captured by Intel RealSense camera. Only 2D print attack was performed with 6 different combinations of used photo state (either full or cut) and 3 different operations like bending the printed paper or movement with different distance from the camera. 
\paragraph{CASIA-SURF CeFA~\cite{db_19_f}} Although CASIA-SURF dataset~\cite{303_19_f} is large-scale, it only includes one ethnicity (Chinese) and one attack type (2D print). So in 2019, CASIA-SURF cross-ethnicity Face Anti-spoofing (CeFA) was released to provide a wider range of ethnicity and attack types, in addition to multiple modalities as CASIA-SURF. The dataset include videos for 1607 subjects, covering three ethnicities (Africa, East Asia, and Central Asia), three modalities, and 4 attack types. Attack types are 2D print attacks, replay attacks in addition to 3D mask and silica gel attacks.

%-------------------------------------------------------------------------
\subsection{Iris spoofing Databases}

%\begin{figure*}[ht]
\begin{figure}[t!]
    \centering
    \begin{subfigure}[t]{0.4\textwidth}
        \centering
        \includegraphics[width=.3\linewidth]{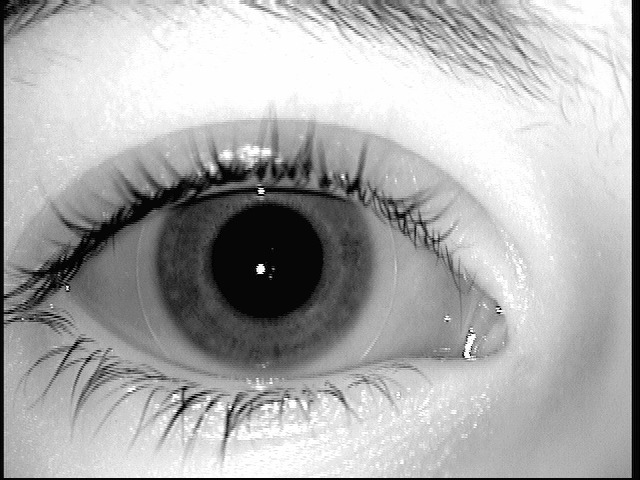}  
        \includegraphics[width=.3\linewidth]{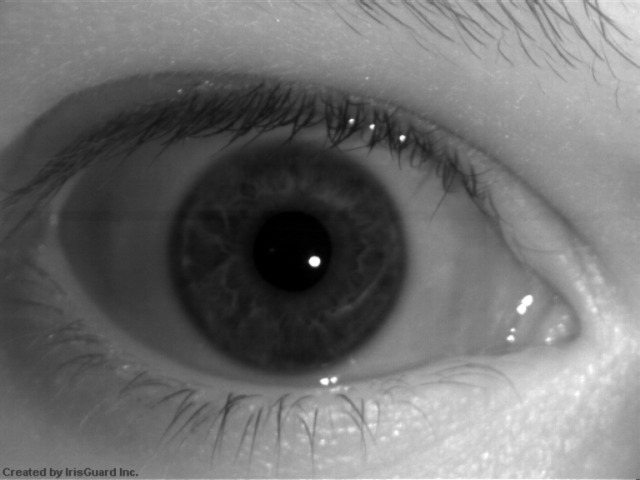}
        \includegraphics[width=.3\linewidth]{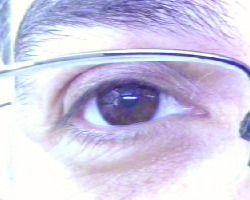}
        \caption{bona-fide}
        \label{fig:iris_real}
    \end{subfigure}
    \hfill
    \begin{subfigure}[t]{0.4\textwidth}
        \centering
        \includegraphics[width=.3\linewidth]{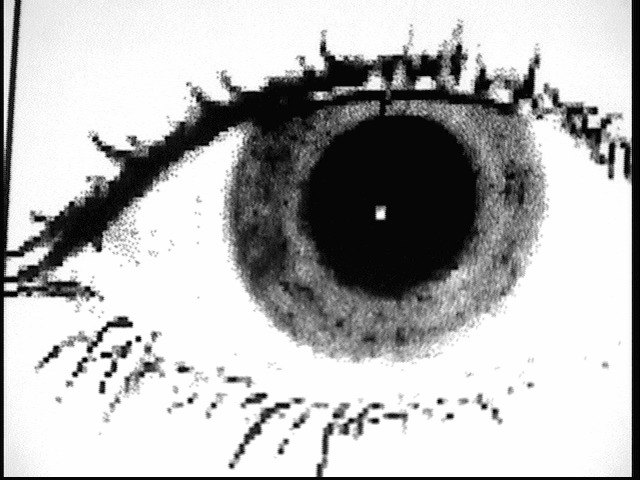}
        \includegraphics[width=.3\linewidth]{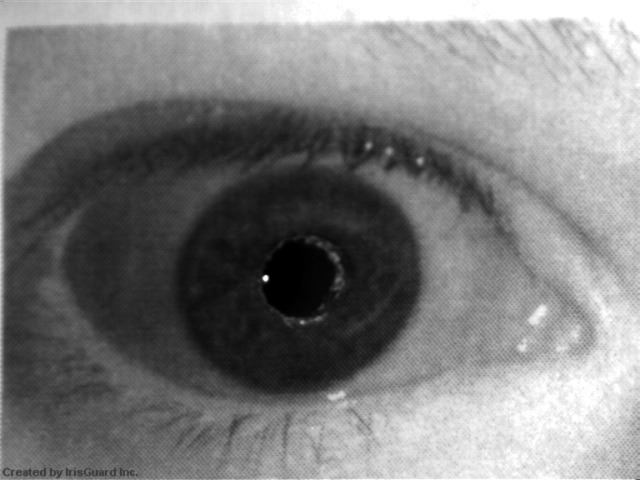}
        \includegraphics[width=.3\linewidth]{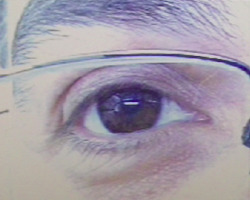}
        \caption{Print attack}
        \label{fig:iris_fake}
    \end{subfigure}
    
    \caption{Sample images from the Iris datasets. Left to right: ATVS-FIr, Warsaw 2013, MobBioFake.
    \figfooter{ATVS-Fir}{id: u0001s0001\_ir\_r\_0002}
	\figfooter{Warsaw}{id: 0039\_R\_1}
	\figfooter{MobBioFake}{id: 007\_R2} }
    \label{fig:iris}
%\end{figure*}
\end{figure}

Many publicly available benchmark datasets were published for the iris presentation attack problem. Some are captured with commercial iris recognition sensors operating with near-infrared illumination like BioSec-ATVS-FIr~\cite{104_12} or LivDet datasets~\cite{109_13, 181_15, 182_17}, while more recent datasets were introduced, captured in the visible light illumination using smartphones, e.g MobBioFake~\cite{155_14, 166_14}. Below are details of the databases used in this paper.

\subsubsection{Biosec ATVS-FIr}

The ATVS-FIr dataset was introduced in 2012~\cite{104_12} from the Biosec dataset~\cite{103_08}. It contains 800 bona-fide and 800 attack photos of 50 different subjects. The attack is performed by enhancing quality of the bona-fide iris photo, printing it on high-quality paper using HP printers, then presenting it to the same LG Iris Access EOU3000 sensor used to capture the bona-fide irises. All images have $640 \times 480$ resolution and are grayscale. For each user 16 different bona-fide photos were captured from which 16 attack photos were generated. Four photos were taken from each of the two eyes in two sessions, so for each user $total = 4~images \times 2~eyes \times 2~sessions = 16$ bona-fide and same for fake attacks. Each eye of the 50 subjects was considered a subject on its own, then the total of 100 subjects were divided into a training set of 25 subjects (200 bona-fide + 200 attack) and a test set of 75 subjects (600 bona-fide + 600 attack)

\subsubsection{LivDet Warsaw 2013}

Warsaw University of Technology published the first LivDet Warsaw Iris dataset in 2013~\cite{109_13} as part of the first international iris liveness competition in 2013~\cite{159_14}. Two more versions of this dataset were later published in 2015~\cite{181_15} and 2017~\cite{182_17}. 
The LivDet-Iris-2013-Warsaw contains 852 bona-fide and 815 attack printout images from 237 subjects. 
For attack images, the real iris was printed using two different printers, then captured by the same IrisGuard AD100 sensor as the bona-fide photos. 

\subsubsection{MobBioFake}
The first competition for iris PAD in mobile scenarios was the Mobi-live competition in 2014~\cite{156_14}. The MobBioFake iris spoofing dataset~\cite{155_14} was generated from the iris images of the MobBIO Multimodal Database~\cite{166_14}. It comprises 800 bona-fide and 800 attack images captured by the back camera (8MP resolution) of an Android mobile smartphone (Asus Transformer Pad TF 300T) in the visible light not in near-infrared like previous iris datasets. The images are color RGB and have a resolution of $250 \times 200$. The MobBioFake database has 100 subjects, each has 8 bona-fide + 8 attack images captured under different illumination and occlusion settings. The attack samples were generated from printed images of the original ones under similar conditions with the same device. The images are split evenly between training and test subsets.

%-------------------------------------------------------------------------
\section{Proposed PAD Approach / Methodology / Algorithm}\label{sec:prop}
Although several previous studies use CNN for presentation attack detection, they either used custom designed networks, e.g. Spoofnet~\cite{308_15_fi, 309_15_i}% in~\cite{308_15_fi, 309_15_i} 
, or used pre-trained CNN to extract features that are later classified with conventional classification algorithms such as SVM. In this paper, we propose to train deeper CNNs for the direct classification of bona-fide and presentation attack images. We choose to assess the state-of-the-art modern deep convolutional neural networks that achieved high accuracies in the task of image classification on ImageNet~\cite{p1_15}. This PAD approach is a passive software texture-based method that does not require additional hardware nor cooperation from the user.

In addition, we compare two different fine-tuning approaches starting from network weights trained on ImageNet classification dataset. A decision can be made to either retrain the whole network weights given the new datasets of the problem, or choose to freeze the weights of most of the network layers and finetune only a set of selected last layers from the network including the final classification layer.

\begin{figure}[t]
\begin{center}

		\includegraphics[width=\linewidth]{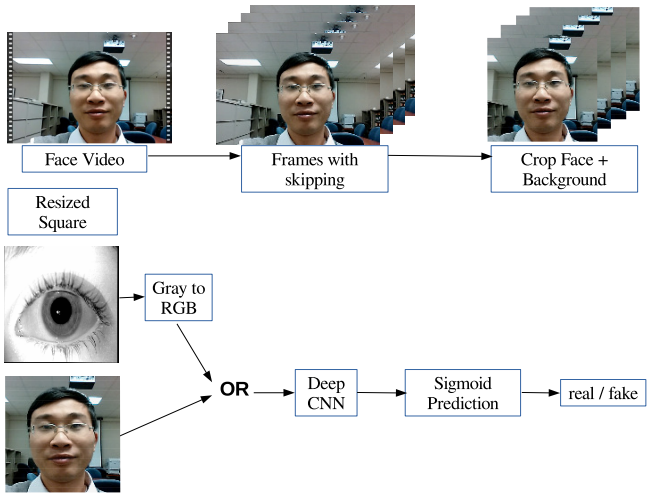}
	\end{center}
\caption{Overview of the PAD algorithm}
\label{fig:f1}
\end{figure}

Figure~\ref{fig:f1} shows the building blocks of the full pipeline proposed to classify face videos or iris images as bona-fide attempts or spoofed attacks. First, the face or iris image is preprocessed and resized, then input to the CNN to a series of convolutional, pooling, and dropout layers. Finally, the last fully connected layer of each network is altered to output a binary bona-fide/attack classification using sigmoid activation instead of the 1000 classes of ImageNet.

%-------------------------------------------------------------------------
\subsection{Preprocessing}
\subsubsection{Face videos}
Each of the face benchmark datasets used in this paper have face coordinates for each frame of each video provided. These coordinates are used to crop a square to be used as input to the network. The square is cropped such that the region of interest (ROI) includes the face and some background in order to account for some contextual information. The side of this square is chosen empirically to be equal twice the minimum of the width and height of the face groundtruth box. The cropped bounding box containing the face is then resized to each network's default size, to be used as input to the first convolutional layer of the CNN.

\subsubsection{Iris images}
When using iris images for recognition or presentation attack detection, most studies in the literature first extract and segment the iris region. However, in this paper we directly use the full iris image captured by the NIR sensor or smartphone camera, in order to make the approach as simple as possible. This helps the the network to learn the best regions, from the full periocular area,  that affect the classification problem. The presence of some contextual information like paper borders can also be useful for the attack detection.

%-------------------------------------------------------------------------
\subsection{Generalization / Data augmentation}
It is known that deep networks have a lot of parameters to train which can cause it to overfit if the training set is not big enough. Data augmentations is a very helpful technique that is used to artificially increase the size of the available training set and to simulate real-life distortions and variations that may not be present in the original set. So during training, for each mini-batch pulled from the training set, images in this batch are randomly cropped, rotated slightly with a random angle, scaled or translated in the x or y axis, so the exact same image is not seen twice by the network. This leads to an artificially simulated larger training dataset and a more generalized network that is robust to changes in pose and illumination.

%-------------------------------------------------------------------------
\subsection{Modern deep CNN architectures}
Based on official results published for recent CNN networks on the ImageNet dataset~\cite{p1_15}, three of the networks that achieved less than 10\% top-5 accuracies are InceptionV3~\cite{p2_15}, ResNet50~\cite{p3_15}, % DenseNet~\cite{p8_16}
and MobileNetV2~\cite{p4_18}. Table~\ref{tabCNNIm} states the single-model top-1 and top-5 error for each of these network on the ImageNet validation dataset, its depth and number of parameters, ordered by ascending top-5 error values. One thing that can be noticed about the higher accuracy networks is the increased number of layers, these networks are much deeper than for example the well-known VGG~\cite{p6_14}.

\begin{table}[t]

\processtable{Performance of selected Deep CNN architectures on ImageNet in terms of Single-model Error \% reported in each model's paper\label{tabCNNIm}}
{\begin{tabular*}{\columnwidth}{L{2.1cm}C{1.5cm}C{1.5cm}cc}

\toprule
&  top-1 (single-crop / multi-crop ) & top-5 (single-crop / multi-crop ) & \# parameters & \# layers \\

\midrule

\textbf{InceptionV3~\cite{p2_15}} & 21.2 / 19.47 (12-crop) & 5.6 / 4.48 (12-crop)
& \textbf{23.9M} & \textbf{159} \\
%\hline
\midrule

MobileNetV2 (alpha=1.4)~\cite{p4_18} & 25.3 / - & $7.58^\star$ / - & 6.2M & 88 \\

\midrule

\textbf{MobileNetV2 (default alpha=1.0)~\cite{p4_18}} &  28 / - & $9.8^\star$  / - & \textbf{3.5M} & \textbf{88} \\

\midrule
VGG16~\cite{p6_14} & - / 24.4 & $9.95^\star$ / 7.1 & 138.4M & 23 \\

\botrule
\end{tabular*}}{${}^\star$ Not available in paper, reported by Keras \url{https://keras.io/}}

\end{table}

For the purpose of this paper, we chose networks that have less than 30 million parameters to finetune. And for the sake of diversity, we 
selected two networks, one with relatively high-number of parameters; InceptionV3~\cite{p2_15} (~24 M parameters with 5.6\% top-5 accuracy), then a more recent and less deep network with far less number of parameters MobilenetV2~\cite{p4_18} (~4 M parameters with 9.9\% top-5 accuracy) which makes it very suitable for mobile applications. Even though all three networks have significantly less parameters than VGG, they are much deeper and have higher accuracy. 

In each architecture, its final prediction softmax layer is removed and the output of the preceding global average pooling layer is fed into a dropout layer (dropout rate of 0.5) then a sigmoid activation layer for final binary prediction.

\begin{table*}[ht]

\fwprocesstable{Number of all parameters in each model and the number of parameters to learn if finetuning only the final block is done.\label{tabFineT}}
		{\begin{tabular*}{\textwidth}{@{\extracolsep{\fill}}ccccc}%{|c|c|C{4cm}|c|c|}
		\toprule

			Model & Total \# of parameters & Last block To train & First trainable layer & \# parameters to learn \\
			\midrule %\hline
			InceptionV3 & 21,804,833 & Block after mixed9 (train mixed9\_1 until mixed10) & conv2d\_90 & 6,075,585 \\

			MobileNetV2 & 2,259,265 & last "\_inverted\_res\_block" & block\_16\_expand & 887,361 \\
			
			%\hline
			\botrule
		\end{tabular*}}{}
\end{table*}

%-------------------------------------------------------------------------
\subsection{Fine-tuning approaches}\label{sec:finetune}
As stated previously, the available datasets are small sized and the used networks have millions of parameters to learn. This will probably lead the model parameters to overfit these small sets if the network weights were trained from scratch; i.e. starting from random weights. The training model will get stuck into a local minima, leading to a poor generalization ability. A better approach is initializing the weights of the network with weights pre-trained on ImageNet, then we have three training options, (1) use transfer learning: feed the images to the network to extract features from the last layer before prediction, then use these features instead of raw image pixels for classification using an SVM or NN, (2) finetune weights of only several last layers of the network while fixing the weights of previous layers to their ImageNet-trained values, or (3) finetune the whole network weights using the new input dataset.

We experiment with only fine-tuning, either the whole network or a few top layers, which we choose to be the last convolutional block. For the approach of fine-tuning the last convolutional block, we state in Table~\ref{tabFineT}, for each architecture, the name of the last block's first layer, and the number of parameters in this block to be finetuned.

%-------------------------------------------------------------------------
\subsection{Single model with multiple biometrics}

In addition to training/testing with datasets that belong to same biometric source, i.e face only or iris only, we experiment with training a single generic presentation attack detection network to detect presentation attack whether the presented biometric is a face or an iris.
We train with different combinations of face+iris sets then cross evaluate the trained models on the other face and iris datasets, to see if this approach is comparable to training on only face images or iris images.

%-------------------------------------------------------------------------
\section{Experiments and Results}\label{sec:exp}

In this section, we describe images and videos preprocessing method used, explain the followed training strategy and experimental setup.

%-------------------------------------------------------------------------
\subsection{Preprocessing}
\subsubsection{Preparing images}
For iris images, only the MobBioFake database consists of color images captured using a smartphone. The other datasets, ATVS-FIr and Warsaw, were obtained with near-infrared sensors producing grayscale images. Since all networks used in this paper were pretrained on colored images of ImageNet, their input layer only accepts color images with three channels and so for grayscale iris images, the single gray channel is replicated two times to form a 3-channel image passed as the network input. The whole iris image is used without cropping nor iris detection and segmentation.

On the other hand, all face PAD datasets consist of color videos. Duration of videos range from 9 to 15 seconds, videos of Replay-Attack database have 25 frames per second (fps), while those of MSU-MFSD have 30 fps. Not all video frames are used, to avoid redundancy and over-fitting to the subjects, so we use frame skipping to sample one frame every 20 frames. This frame dropping leads to sampling around 10 to 19 frames per video based on its length. For each sampled frame, instead of using the whole frame as the network input, it is first cropped to either include only the face region or a box that has approximately twice the face area to include some background context. Annotated face bounding boxes have a side length ranging from $70px$ to $100px$ in Replay-Attack videos, and from $170px$ to $250px$ for MSU-MFSD.

Table~\ref{tabDB2} shows the training, validation and testing sizes (in number of images) for each of the datasets used in the experiments (after frame dropping in videos). Emphasized in bold is the total number of images available for training the CNN for each database.

\begin{table}%[!b]
\processtable{Number of images used in training and testing. For face datasets, number of videos is shown followed by number of frames between (). For iris datasets, only number of dataset images is mentioned.\label{tabDB2}}
{\begin{tabular*}{20pc}{@{\extracolsep{\fill}}lcr@{}}
\toprule

\multirow{2}{*}{Database} & \multirow{2}{*}{Subset} & \multicolumn{1}{c}{Number of videos (frames$^\star$)} \\
 & & \multicolumn{1}{c}{Attack + bona-fide}  \\

\midrule
\multirow{3}{*}{Replay-Attack} & train &  $300 (3414) + 60 (1139) = \textbf{\emph{4553}}$ \\
 & devel & $300 (3411) + 60 (1140)$ \\
 & test & $400 (4516) + 80 (1507)$  \\ 

\midrule
\multirow{2}{*}{MSU-MFSD} & train &  $90 (1257) + 30 (419) = \textbf{\emph{1676}}$ \\
 & test & $120 (1655) + 84 (551)$  \\

\midrule
\multirow{3}{*}{Replay-Mobile} & train &  $192 (2851) + 120 (1762) = \textbf{\emph{4613}}$ \\
& devel & $256 (3804) + 160 (2356)$ \\
& test & $192 (2842) + 110 (1615)$  \\ 

\midrule
\multirow{2}{*}{ATVS-FIr} & train & $200 + 200 = \textbf{\emph{400}}$  \\
& test & $600 + 600$ \\ 

\midrule
\multirow{2}{*}{Warsaw 2013} & train & $203 + 228 = \textbf{\emph{431}}$  \\
& test & $612 + 624$ \\

\midrule
\multirow{2}{*}{MobBioFake} & train & $400 + 400 = \textbf{\emph{800}}$ \\
& test & $400 + 400$ \\ 

\botrule
\end{tabular*}}{$\star$ (frames: skipping every 20 frames)}

\end{table}

The images are then resized according to the default input size values for each of the used networks. That is $224 \times 224$ for MobilenetV2, and $299 \times 299$ for InceptionV3. 
The RGB values of input images to the InceptionV3 and the MobilenetV2 networks are first scaled to have a value between -1 and 1.

\subsubsection{Data augmentation}
As shown in Table~\ref{tabDB2}, the total number of images available from each dataset to train the CNN is relatively much smaller than sizes of datasets usually used for deep networks training. This could cause the network to memorize these small set of samples instead of learning useful features and so fail to generalize. So in order to reduce overfitting, data augmentation is used during training. Images are randomly transformed before feeding to the network with one or more of the following transformations: Horizontal flip, Rotation between 0 and 10 degrees, Horizontal translation by 0 to 20 percent of image width, Vertical translation by 0 to 20 percent of image height, or Zooming by 0 to 20 percent

%-------------------------------------------------------------------------
\subsection{Training strategy and Hyperparameters}
For training the CNN networks, we used Adam optimizer~\cite{adam} with $beta1=0.9$, $beta2=0.999$, and learning rate $1 \times 10^{-4}$.% or $1 \times 10^{-5}$. 
We trained the network for max of 50 epochs but added a stopping criteria if the training or validation accuracy reached 99.95\% or validation accuracy did not improve (with $delta = 0.0005$) for 40 epochs. %Mini-batch size was set to 16. 

Intermediate models with high validation accuracy were saved and finally the model with the least validation error was chosen for inference. Because of the randomness introduced by the data augmentation and the dropout, several training sessions may produce slightly different results, specially in cross-validation, for this reason, three runs were performed for each configuration (model + dataset + 
 fine-tuning approach) and the min error is reported. 
 
\begin{table*}[h] 
\fwprocesstable{Intra and cross-dataset results. Test HTER reported with Replay-Attack and Replay-Mobile testing, otherwise Test EER.\label{tabResAB}}
		{\begin{tabular*}{\textwidth}{@{\extracolsep{\fill}}llcccc}
		\toprule

			\multirow{2}{*}{Train Database} & \multirow{2}{*}{Test Database} & \multicolumn{2}{c}{Finetune Approach (A)}  & \multicolumn{2}{c}{Finetune Approach (B)} \\

			&  & InceptionV3 & MobilenetV2 & InceptionV3 & MobilenetV2 \\
%			\hline
%			\hline
			\midrule
			
			\multirow{3}{*}{Replay-Attack} & Replay-Attack 
			& 6.9\%	& 1.6\% 
			& \textbf{\underline{0\%}}	& 0.63\%
			\\
			%\cline{2-6}
			& MSU-MFSD 
			& 33\%	& 30\% 
			& \underline{\textbf{17.5\%}}	& 22.5\%
			\\
			%\cline{2-6}
			& Replay-Mobile 
			& \textbf{35.8\%} & 43.4\%  
			& 64.7\% &  56.7\% 
			\\

			\midrule
			
			\multirow{3}{*}{MSU-MFSD} 
			
			& Replay-Attack 
			& 34.7\% & 22.9\%
			& 23.8\% & \textbf{\underline{13.7\%}} 
			\\
%			\cline{2-6}
			& MSU-MFSD 
			& 7.5\% &	2.05\% 
			& \underline{\textbf{0\%}} &	2.08\% 
			\\ 
%			\cline{2-6}
			& Replay-Mobile 
			& 32.6\% & 26.8\% 
			& 24.6\% & \textbf{\underline{23.7\%}}
			\\

			\midrule
						
			\multirow{3}{*}{Replay-Mobile} 
						& Replay-Attack 
			& 33.2\% & 45.1\%
			& \textbf{31.1\%}	& 45.5\% 
			 \\ 
			%\cline{2-6}
			& MSU-MFSD 
			& 35.4\% & \textbf{30\%} 
			& 32.9\% & 37.5\%
			\\ 

			& Replay-Mobile 
			& 3.6\% & 3.2\% 
			& \textbf{\underline{0\%}} & \textbf{\underline{0\%}} 
			\\ 

			\midrule
			
			\multirow{2}{*}{ATVS-FIr} & ATVS-FIr 
			& \textbf{\underline{0\%}} & \textbf{\underline{0\%}} 
			& \textbf{\underline{0\%}} & \textbf{\underline{0\%}} 
			\\
			& Warsaw 2013 
			& 1.9\%  & 5.4\% 
			& \underline{\textbf{0.16\% }} &	1.6\%
			\\

& MobBioFake (Grayscale) 
		& 44\%  & 39\% 
		& \textbf{36.5\%} &	39\%
			\\
			\midrule
			
			\multirow{2}{*}{Warsaw 2013} 
			& ATVS-FIr 
			& \textbf{\underline{0\%}} & 1\% 
			& \textbf{\underline{0.17\%}} & 0.5\%  
			\\
			& Warsaw 2013 
			& \textbf{\underline{0\%}} & 0.32\% 
			& \textbf{\underline{0\%}} & \textbf{\underline{0\%}} 
			\\
			& MobBioFake (Grayscale) 
			& 43\%  & 39\% 
			& \underline{\textbf{32.8\% }} &	47\%
			\\

			\midrule
			
\multirow{3}{*}{MobBioFake}
			& ATVS-FIr 
			& \textbf{10\%}  & 30\% 
			& 18.7\% & 20.2\%
			\\
			& Warsaw 2013 
			& \textbf{15.7\%}  & 32\%
			& 23.4\% & 18.8\%
			 \\
			& MobBioFake 
			& 8.75\% & 7\% 
			& \textbf{\underline{0.5\%}} & 0.75\% 
			\\
			\midrule
			
			\multirow{3}{*}{MobBioFake (Grayscale)} 
			& ATVS-FIr 
			& \textbf{2.5\%}  & 23.7\%
			& 12\% & 16.2\% 
			\\
			& Warsaw 2013 
			& \textbf{6.2\%}  & 17.3\% 
			& 10.1\% & 17.2\%
			\\
%			\cline{2-6}
			& MobBioFake (Grayscale) 
			& 10.5\%  & 7.3\%
			& \textbf{\underline{0.25\%}} & 1.25\%
			 \\
			\botrule

		\end{tabular*}}{}
\end{table*}
%\end{table}

%-------------------------------------------------------------------------
\subsection{Evaluation}
\subsubsection{Videos scoring}
The CNN accepts images as input, so we treated each frame in the face video as a separate image. Then at reporting the final score for each video we perform score-level fusion by averaging scores of the samples from the same video. Accuracy and errors can then be calculated based on these video-level scores not frame-level scores as some reported results in literature.

\subsubsection{Evaluation metrics}
The equal-error-rate (EER) is used to report the error of the test set. EER is defined as the rate at which both the False Rejection Rate (FRR) and the False Acceptance Rate (FAR) are equal. FRR is the probability of the algorithm rejecting bona-fide samples as attack ones, which is equal to $FR / total~bona~fide~attempts$, while FAR is the probability by which the algorithm might accept attack samples and consider them bona-fide, $FA / total~attack~attempts$.

For datasets that have a separate development subset; e.g. Replay-Attack, first a score threshold is calculated that achieves EER on the devel set. Then this threshold is used to calculate error in the test set known as half-total error rate (HTER) which is equal to the summation of the False Rejection Rate (FRR) and the False Acceptance Rate (FAR) at a threshold, divided by two, $HTER = (FRR+FAR) / 2$.

An ISO/IEC international standard for PAD testing was described in ISO/IEC 30107-3:2017~\footnote{\url{https://www.iso.org/obp/ui/##iso:std:iso-iec:30107:-3:ed-1:v1:en}} where PAD subsystems are to be evaluated using two metrics; attack presentation classification error rate (APCER) and bona-fide presentation classification error rate (BPCER). However, we report results in this paper using HTER and EER to be able to compare with state-of-the-art published results, EER can also be understood as the rate at which APCER is equal to BPCER.

\subsubsection{Cross-dataset evaluation}
In order to test the generalization ability of the used networks, we perform cross-dataset evaluation where we train on a dataset and test on another dataset. 
For the iris datasets, we know of no previous work to have performed EER cross-dataset evaluation on iris PAD databases before. 

%-------------------------------------------------------------------------
\subsection{Experimental setup}
For implementation we used Keras \footnote{\url{https://github.com/keras-team/keras}} with tensorflow \footnote{\url{https://www.tensorflow.org/}} backend for the deep CNN models, and for datasets management and experiments pipelines we used Bob package~\cite{bob2012,bob2017}. 
~Experiments were performed on NVIDIA GeForce 840m GPU with CUDA version 9.0.

% --------------------------------------------------------------

\begin{table*}[h] %t] %[hb]
\fwprocesstable{Training with Face+Iris images. Intra and cross-dataset results. Test HTER reported with Replay-Attack and Replay-Mobile testing, otherwise Test EER. \label{tabResFaceIris}}
		{\begin{tabular*}{\textwidth}{@{\extracolsep{\fill}}l|l|cc|cc|cc|cc}
		\toprule

		& Train Iris Database: & \multicolumn{2}{c}{ATVS-FIr} & \multicolumn{2}{c}{Warsaw 2013} & \multicolumn{2}{c}{MobBioFake (RGB)} & \multicolumn{2}{c}{MobBioFake (Grayscale)}\\

\midrule
			\multirow{1}{1.75cm}{Train Face Database}  & Test Database & InceptionV3 & MobilenetV2 & InceptionV3 & MobilenetV2 & InceptionV3 & MobilenetV2 & InceptionV3 & MobilenetV2\\
			& & & & & & & & & \\
			\midrule
			
			\multirow{6}{*}{Replay-Attack}  
			& \multirow{2}{2.7cm}{-- Number of training epochs} & 2 & 6 & 3 & 6 & 4 & 9 & 2 & 8 \\
			& & & & & & & & & \\
			& \textit{Replay-attack} & 0\%	& 0.63\%	& 0\%	& 0\%	& 0.25\%	& 1.5\% & 0\% & 1.4\% \\
			& MSU-MFSD & 29.6\% & 20.4\% &		\textbf{\underline{15\%}} &	20\% &	19.6\% &	25\%\ & 30\% & 17.5\% \\
			%\cline{2-6}
			& Replay-Mobile & 58.1\% &	48.7\% &		57.3\% &	\textbf{35.4\%} &	60.6\% &	58.1\% & 39\% & 40\% \\
			& ATVS-FIr	&  0\% &	0\% &	10.5\% &	\textbf{2.7\%} &	5.17\% &	8.8\% & 3.5\% & 12.3\% \\
		 & Warsaw 2013 &  \textbf{\underline{0.16\%}} &	\textbf{\underline{0.16\%}} &		0\% &	0\% &	4.13\% &	1.62\% & 15.2\% & 5.7\% \\
			& MobBioFake (RGB) & 58.5\% &	71.3\% &		56\% &	\textbf{45.5\%} &	0.75\% &	0.25\% & 1.75\% & 2\% \\

		& MobBioFake (Grayscale) & 50\% & 69\% & 56.3\% & \textbf{38.5\%} & 1.5\% & 0.75\% & 1.25\% & 1\% \\

		\midrule
		
		\multirow{6}{*}{MSU-MFSD}  
		& \multirow{2}{2.7cm}{-- Number of training epochs} & 4 &	6	&	3	& 5 &	10 &	9 & 7 & 9 \\
			& & & & & & & & & \\
			& Replay-Attack & 22.8\%  &	14.5\%  &		25.3\%  &	\textbf{\underline{11.9\%}}  &	22.8\%  &	27.7\% & 16.6\% & 21.6\% \\
			%\cline{2-6}
			& \textit{MSU-MFSD} & 0\%  &	2.08\%  &		0.4\%  &	2.08\%  &	2.5\%  &	2.08\% & 0.4\% & 2.5\% \\
			%\cline{2-6}
			& Replay-Mobile & 28.7\%  &	26.1\%  &		19.2\%  &	24.8\%  &	22.7\%  &	26.8\% & 25.2\% & \textbf{\underline{17\%}} \\
			& ATVS-FIr	& 0\%  &	0\%  &		\textbf{\underline{0.33\%}}  &	2.3\%  &	11.8\%  &	12.5\% & 0.8\% & 12.33\% \\
		 & Warsaw 2013 & \textbf{0.6\%}  &	0.8\%  &		0\%  &	0\%  &	43.9\%  &	6.55\% & 7.4\% & 9.1\% \\
			& MobBioFake (RGB) & 52.5\%  &	\textbf{\underline{39.2\%}}  &		45.5\%  &	42\%  &	2\%  &	1.25\% & 2.5\% & 2.5\% \\
			& MobBioFake (Grayscale) & 45.3\% & 38.5\% & \textbf{34.5\%} & 43\% & 22\% & 4.25\% & 1.75\% & 2\% \\

		\midrule		
		\multirow{6}{*}{Replay-Mobile}  
		& \multirow{2}{2.7cm}{-- Number of training epochs} & 2 &	3	& 1	 & 3 &	4	& 7 & 5 & 4 \\
			& & & & & & & & & \\
			& Replay-Attack & \textbf{24.2\%} &	42.6\% &		31\% &	35.7\% &	41.6\% &	64.3\% & 33.7\% & 54.6\% \\
			%\cline{2-6}
			& MSU-MFSD & 32.5\% &	\textbf{30.4\%} &		32.5\% &	42.5\% &	60\% &	52.5\% & 40\% & 44.6\% \\
			%\cline{2-6}
			& \textit{Replay-Mobile} & 0\% &	0\% &		0\% &	0\% &	0\% &	0.26\% & 0\% & 0.46\% \\
			& ATVS-FIr	& 0\% &	0\% &		\textbf{0.83\%} &	6\% &	15.5\% &	22.5\% & 6.5\% & 28.2\% \\
		 & Warsaw 2013 & \textbf{2.1\%} &	4\% &		0\% &	0\% &	9.6\% &	28.9\% & 22.1\% & 26.1\% \\
			& MobBioFake (RGB) & \textbf{46.5\%} &	53\% &		49.3\% &	54.3\% &	1\% &	0.75\% & 8.8\% & 3.25\% \\
		& MobBioFake (Grayscale) & \textbf{\underline{33.5\%}} & 43.3\% & 41.7\% & 45.8\% & 5.3\% & 6.5\% & 2\% & 1.75\% \\
			\botrule
		\end{tabular*}}{(\textbf{\underline{bold-underlined}}): best cross-dataset error for each test dataset, (\textbf{bold}): best cross-dataset error for each test dataset in each row (i.e: given a certain face train set)}%Table footnote}

\end{table*}

%-------------------------------------------------------------------------
%-------------------------------------------------------------------------

\begin{table*}[ht] %[hb]
\fwprocesstable{Best cross-dataset results and comparison with SOTA. Test HTER reported with Replay-Attack testing, otherwise Test EER. Underlined is the best cross-dataset error for the test dataset.\label{tabSOTA2}}
{\begin{tabular*}{\textwidth}{@{\extracolsep{\fill}}llccc}
		\toprule
		
		Train Database & Test Database & SOTA / corresponding intra-dataset error & \multicolumn{2}{c}{Our Best / corresponding intra-dataset error} \\
		 & &  & Model trained with only 1 biometric & Model trained with face + iris \\
\midrule

\multirow{2}{*}{Replay-Attack} & \multirow{2}{*}{MSU-MFSD} & 28.5\% / 8.25\%& 17.5\% / 0\%	&  \underline{\textbf{15\% / 0\%}} \\
&  & GuidedScale-LBP~\cite{560_18_f} "LGBP"  & "InceptionV3 (B)" & "InceptionV3" (Warsaw 2013)\\
& \multirow{2}{*}{Replay-Mobile} & 49.2\% / 3.13\% & 35.8\% / 6.9\% &  \underline{\textbf{35.4\% / 0\%}} \\
& & GuidedScale-LBP~\cite{560_18_f} "LBP+ GS-LBP" & "InceptionV3 (A)" & "MobilenetV2" (Warsaw 2013)\\
\midrule

\multirow{4}{*}{MSU-MFSD} & \multirow{2}{*}{Replay-Attack} & 21.40\% / 1.5\% & 13.7\% / 2.08\%  & \textbf{\underline{11.9\%}} / 2.08\% \\
& & Color-texture~\cite{555_18_f} & "MobilenetV2 (B)" & "MobilenetV2" (Warsaw 2013) \\
& \multirow{2}{*}{Replay-Mobile} & 32.1\% / 8.5\% & 23.7\% / 2.08\% &   \textbf{\underline{17\% / 2.5\%}} \\
& & GuidedScale-LBP~\cite{560_18_f} "LBP+ GS-LBP" & "MobilenetV2 (B)" &  "MobilenetV2" (MobBioFake (Gray))\\
\midrule

\multirow{4}{*}{Replay-Mobile} & \multirow{2}{*}{Replay-Attack} & 46.19\% / 0.52\% &  31.1\% / 0\%  & \textbf{24.2\% / 0\%} \\
& & GuidedScale-LBP~\cite{560_18_f} "LGBP" &  "InceptionV3 (B)" & "InceptionV3" (ATVS-FIr) \\
%\cline{2-4}
& \multirow{2}{*}{MSU-MFSD} & 33.56\% / 0.98\% & 30\% / 3.2\%	& \textbf{30.4\% / 0\%} \\
& & GuidedScale-LBP~\cite{560_18_f} "LBP+ GS-LBP" & "MobilenetV2 (A)" & "MobilenetV2" (ATVS-FIr)\\
%\hline
\midrule

\multirow{4}{*}{ATVS-FIr} 
& \multirow{2}{*}{Warsaw 2013} & None available  & \underline{\textbf{0.16\% / 0\%}} &\underline{\textbf{0.16\% / 0\%}} 	\\
& & & "InceptionV3 (B)" & "Any" (Replay-Attack)\\
& \multirow{2}{*}{MobBioFake (Gray)} & None available & 36.5\% / 0\%	& \underline{\textbf{33.5\% / 0\%}} \\
& & & "InceptionV3 (B)" & "InceptionV3" (Replay-Mobile)\\
%\hline
\midrule

\multirow{4}{*}{Warsaw 2013} &  \multirow{2}{*}{ATVS-FIr} & None available & \textbf{\underline{0\% / 0\%}} & 0.33\% / 0\%\\
& &  & "InceptionV3 (A)" & "InceptionV3" (MSU-MFSD) \\
& \multirow{2}{*}{MobBioFake (Gray)} & None available & \textbf{\underline{32.8\% / 0\%}}	& 34.5\% / 0\% \\
& & & "InceptionV3 (B)" & "InceptionV3" (MSU-MFSD)\\
%\hline

\midrule
\multirow{4}{*}{MobBioFake (Gray)} 
&  \multirow{2}{*}{ATVS-FIr} & None available & 2.5\% / 10.5\% & \textbf{0.8\% / 1.75\%} \\
& & & "InceptionV3 (A)" & "InceptionV3" (MSU-MFSD) \\
&  \multirow{2}{*}{Warsaw 2013} & None available & 6.2\% / 10.5\% & {\textbf{5.7\% / 1\%}} \\
& & & "InceptionV3 (A)" & "MobilenetV2" (Replay-Attack) \\
\botrule

\end{tabular*}}{}
\end{table*}

%-------------------------------------------------------------------------
%-------------------------------------------------------------------------

\subsection{Experiment 1 Results}\label{sec:res1}
In the first experiment, we compare the two different finetuning approaches explained in Section~\ref{sec:finetune}. For this experiment, training and testing was done using face only or iris only datasets. We performed intra and cross-dataset testing to evaluate the effectiveness and generalization of the proposed approach, and used benchmark datasets to be able to compare against state-of-the-art. For each finetuning approach, we performed three runs and reported the minimum error obtained. 
In Table~\ref{tabResAB} we refer to the first finetuning approach as (A) where the weights of only the last block is finetuned, while approach (B) is when all the layers' weights are finetuned using the given dataset.

It can be noticed from Table~\ref{tabResAB} that finetuning all the network weights achieved better results than fixing weights of first layers to their ImageNet-trained values and only finetuning just the final few layers. For intra-datasets error, we achieved State-of-the-art (SOTA) 0\% test error for all datasets using InceptionV3, while MobilenetV2 achieved 0\% for most datasets and 1-2\% on MSU-MFSD and MobBioFake. 
The table also shows that cross-dataset errors on the iris datasets ATVS-FIr and Warsaw are much lower when grayscale version of the MobBioFake iris dataset was used for training instead of using the original color images of the dataset, without much affecting the intra-dataset error on MobBioFake itself.

\subsection{Experiment 2 Results}\label{sec:res2}
For the second experiment, we only use finetuning approach (B); adjusting the weights of all the network's layers using the training images. Here we use both face and iris images as train images input to a single model for classifying bona-fide from attack presentation in general regardless the biometric type. 
 
Results in Table~\ref{tabResFaceIris} show that using both face and iris images in training a single model to differentiate between bona-fide and presentation attack images achieved very good results on both face and iris test sets. For some cases in face datasets, this combination boosted the performance for cross-dataset testing where for example on Replay-Attack test dataset 11.9\% HTER was achieved when training a MobilenetV2 using MSU-MFSD and Warsaw Iris dataset, vs 13.7\% HTER when only MSU-MFSD was used for training as reported in Table~\ref{tabResAB}.

Table~\ref{tabResFaceIris} also confirms findings from Table~\ref{tabResAB}; that training with grayscale version of the MobBioFake iris dataset gives better cross-dataset error on the other gray iris datasets, without much affecting the intra-dataset error on the MobBioFake dataset itself. Not only this, but it also achieves lower cross-dataset error rates when evaluating on face datasets, by helping the network to focus on color features when only testing is performed on a face image, since all training face images were colored but all iris training images were grayscale.

%-------------------------------------------------------------------------
%-------------------------------------------------------------------------

\begin{figure}[!] %[t!]
    \centering
        \includegraphics[width=.19\linewidth]{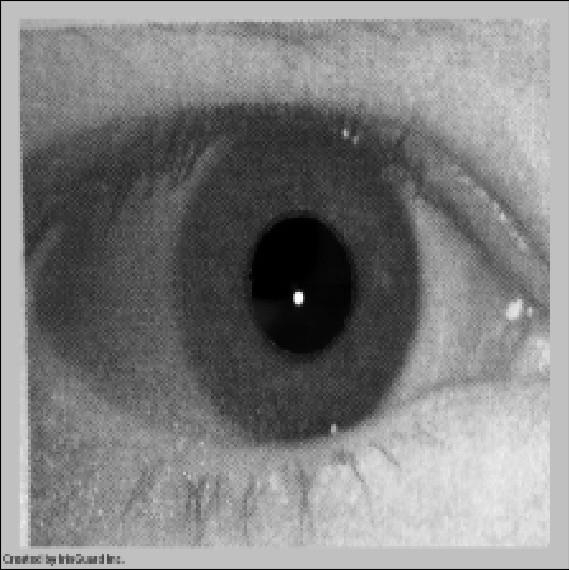}
        \includegraphics[width=.19\linewidth]{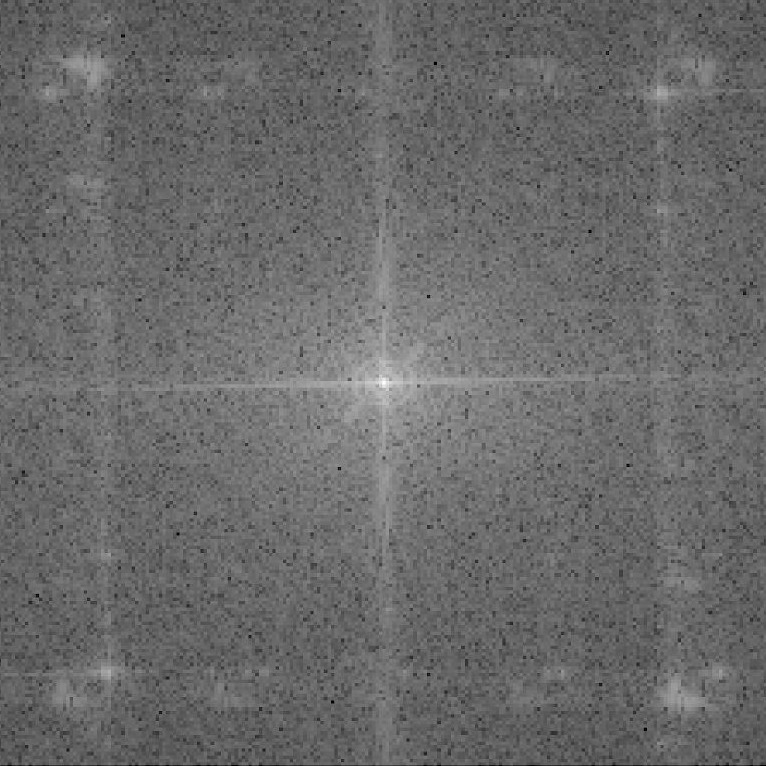}
        \includegraphics[width=.19\linewidth]{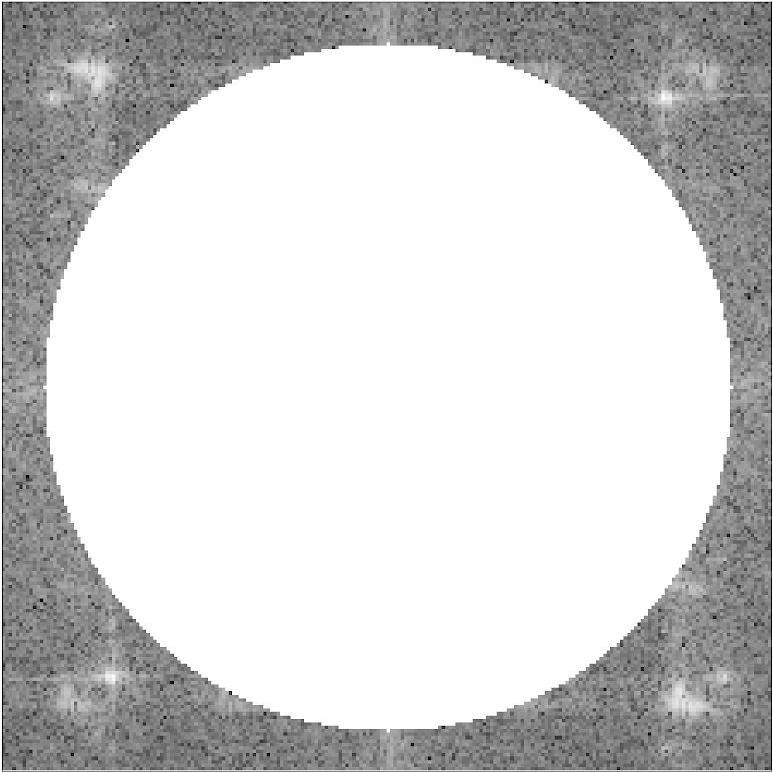}
        \includegraphics[width=.19\linewidth]{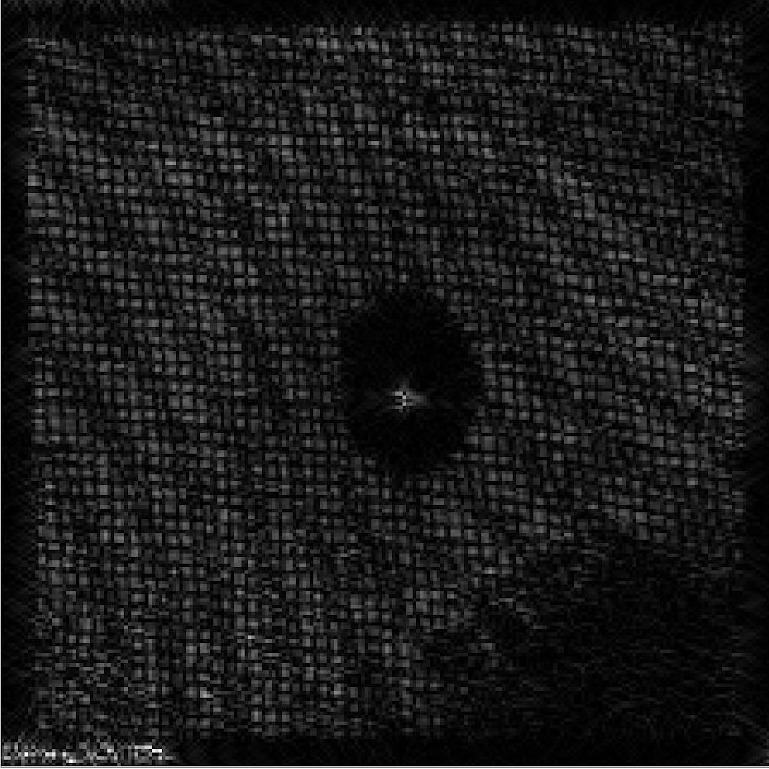}
        \includegraphics[width=.19\linewidth]{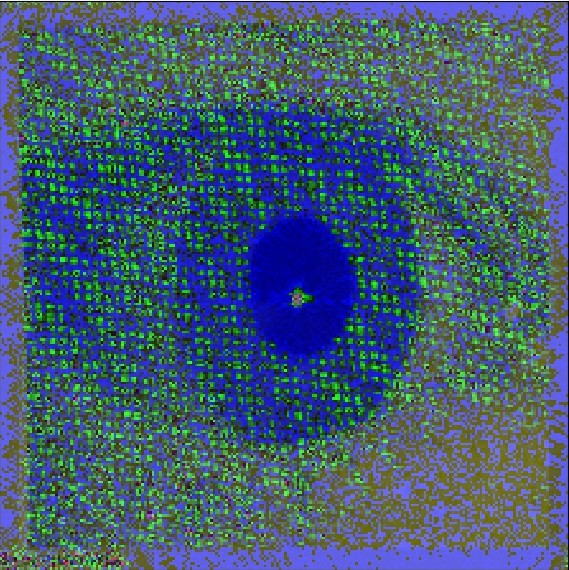}

    \caption{High frequency components Attack iris images from Warsaw dataset.
     \figfooter{*}{Left-to-right: Input image, Frequency compoents (Fourier), High-pass-filter, Inverse fourier on image, High frequency components overlayed over image}
}
    \label{fig:analysis_hpf}
\end{figure}

%-------------------------------------------------------------------------
%-------------------------------------------------------------------------

\begin{figure}[!] %[t!]
    \centering
    \begin{subfigure}[t]{0.5\textwidth}
        \centering
        \includegraphics[width=.23\linewidth]{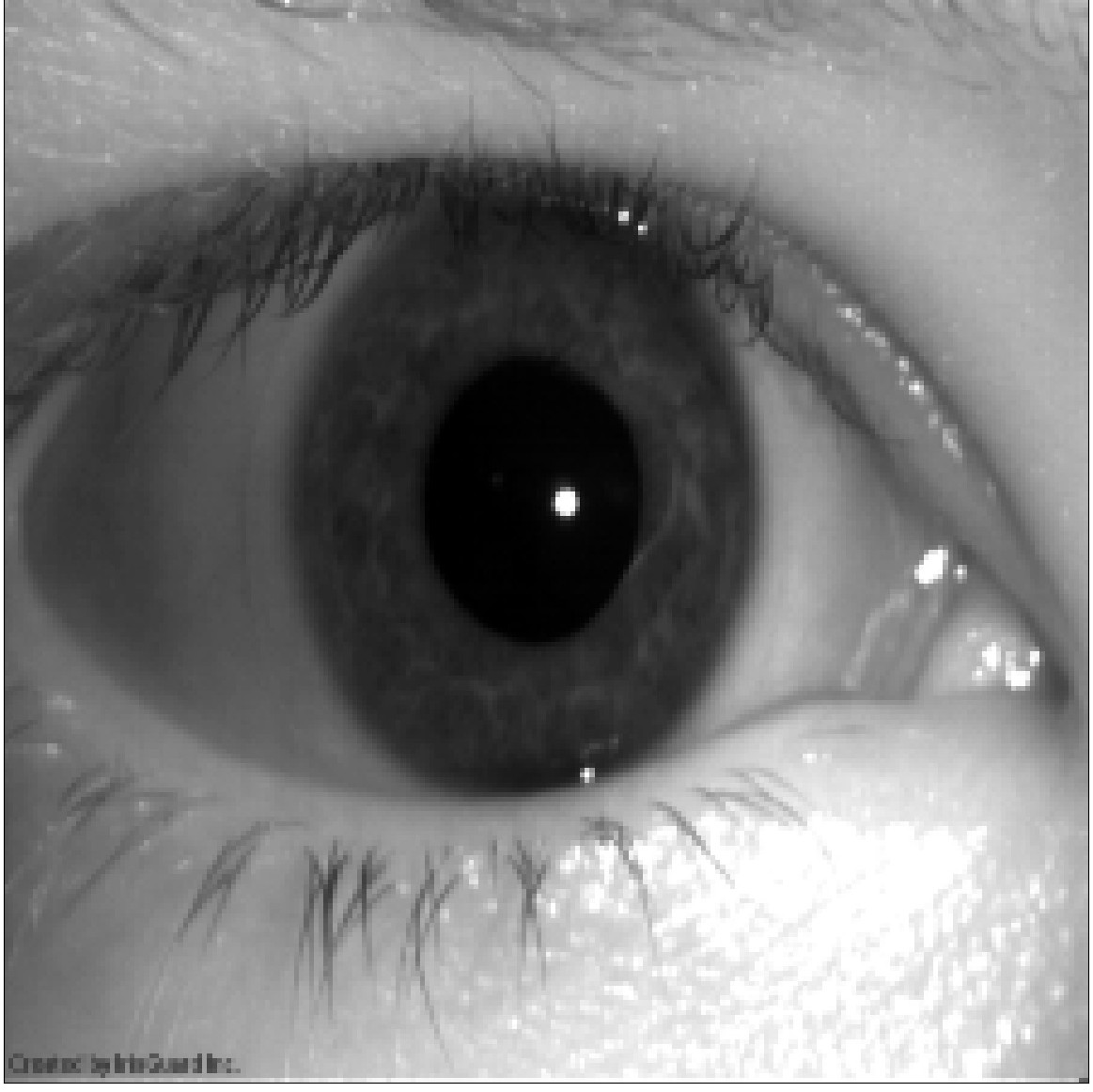}
        \includegraphics[width=.23\linewidth]{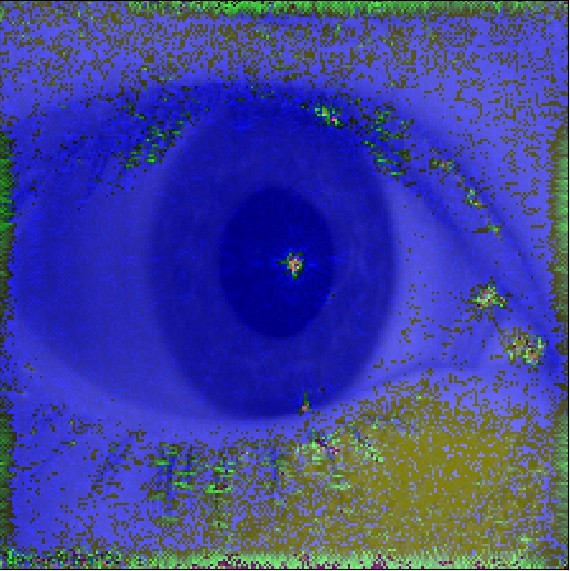}
        \includegraphics[width=.23\linewidth]{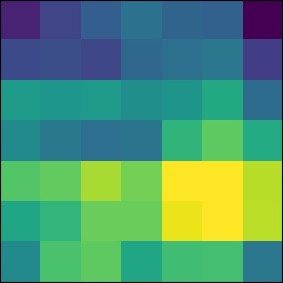}
        \includegraphics[width=.23\linewidth]{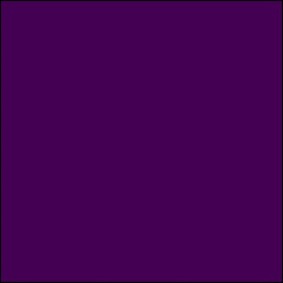}\\
        \includegraphics[width=.23\linewidth]{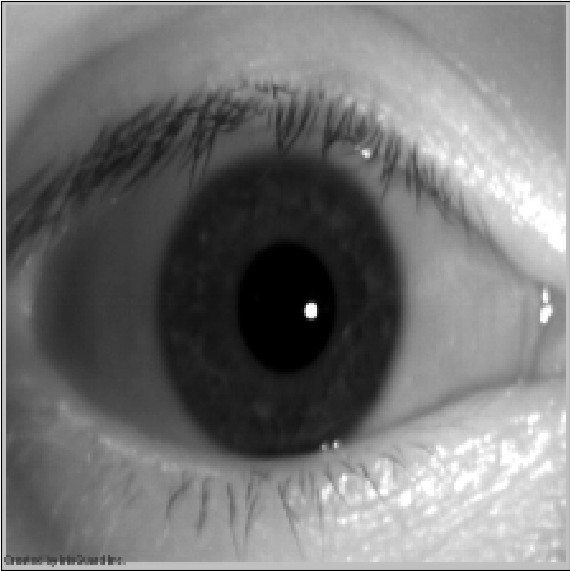}
        \includegraphics[width=.23\linewidth]{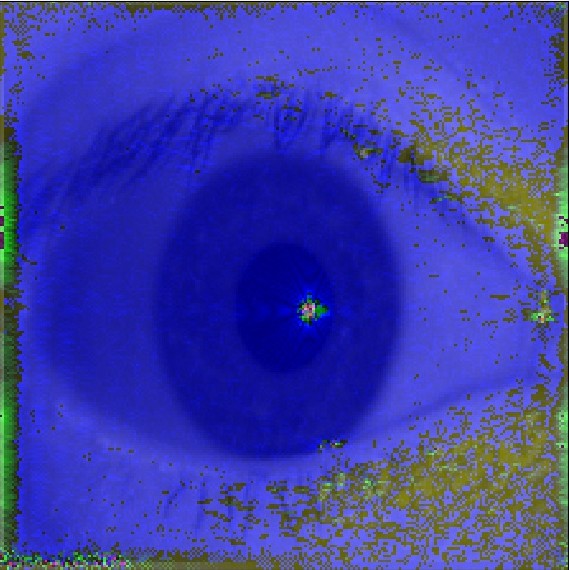}
        \includegraphics[width=.23\linewidth]{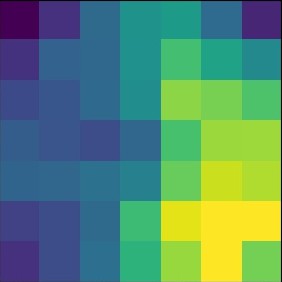}
        \includegraphics[width=.23\linewidth]{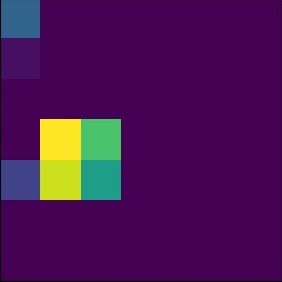}
        
        \caption{Bona-fide Iris samples}
        \label{fig:freqRealIris}
    \end{subfigure}%
    \hfill
    \begin{subfigure}[t]{0.5\textwidth}
        \centering

        \includegraphics[width=.23\linewidth]{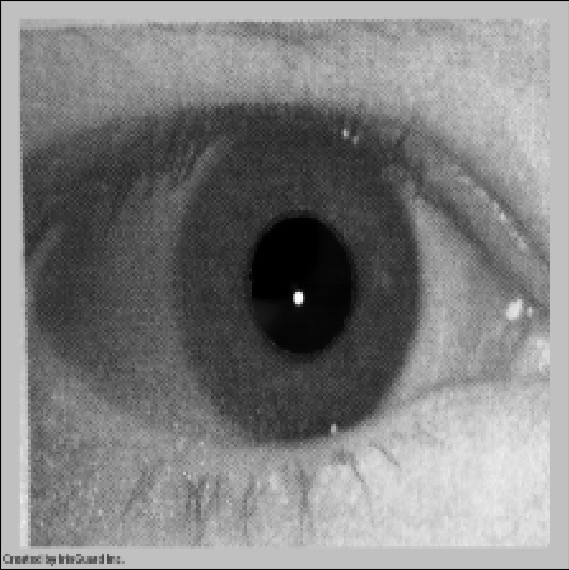}
        \includegraphics[width=.23\linewidth]{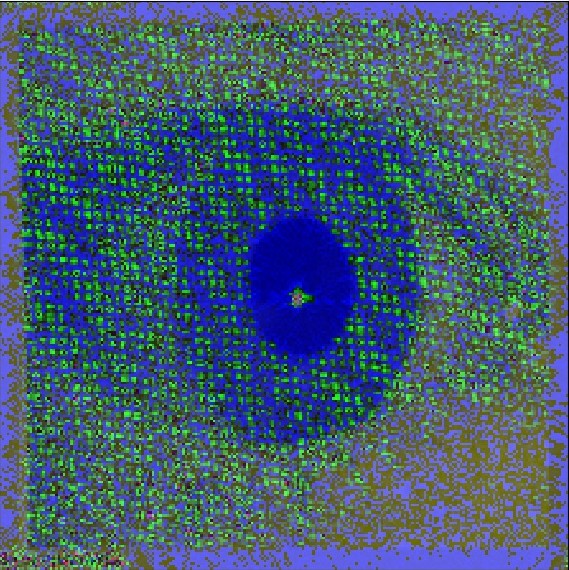}
		\includegraphics[width=.23\linewidth]{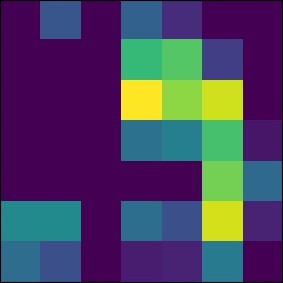}
		\includegraphics[width=.23\linewidth]{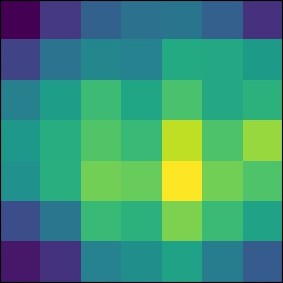} \\
        \includegraphics[width=.23\linewidth]{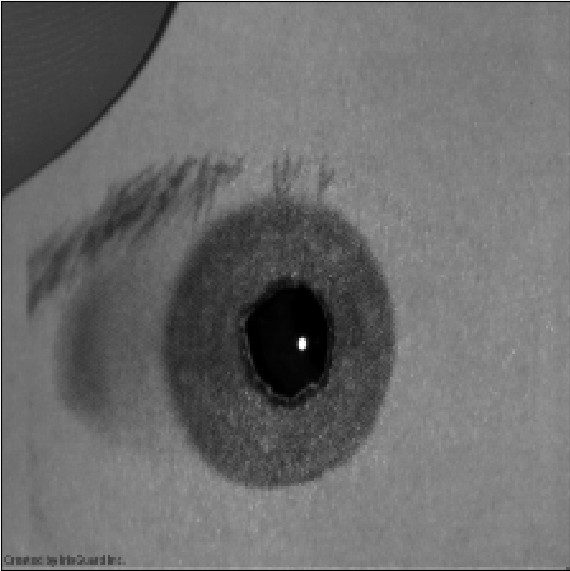}
        \includegraphics[width=.23\linewidth]{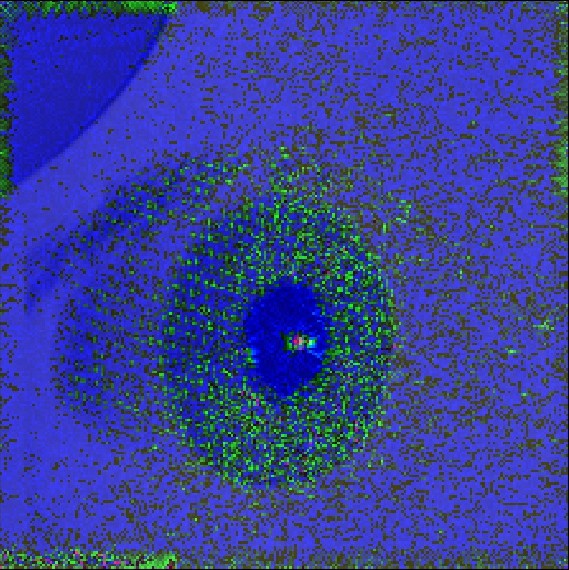}
        \includegraphics[width=.23\linewidth]{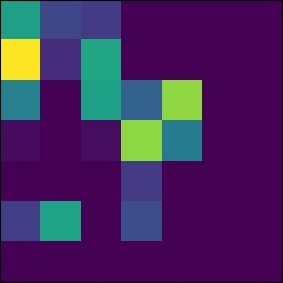}
        \includegraphics[width=.23\linewidth]{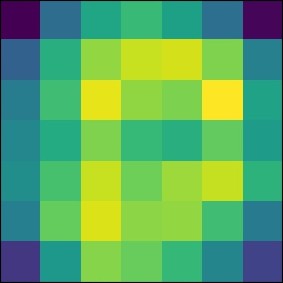}
        
        \caption{Attack Iris samples}
        \label{fig:freqAttackIris}
    \end{subfigure}

    \caption{High frequency components in bona-fide vs Attack iris images from Warsaw datase.
     \figfooter{*}{Left-to-right: Input image, High frequency components overlayed over image, Network activation response of realF, Network activation response of attackF}
     \figfooter{*}{Top: 0050\_R\_2, Bottom: 0088\_R\_3}
}
    \label{fig:analysis_iris}
\end{figure}

%-------------------------------------------------------------------------
%-------------------------------------------------------------------------

\begin{figure}[!] %[t!]
    \centering
    \begin{subfigure}[t]{0.5\textwidth}
        \centering
        \includegraphics[width=.23\linewidth]{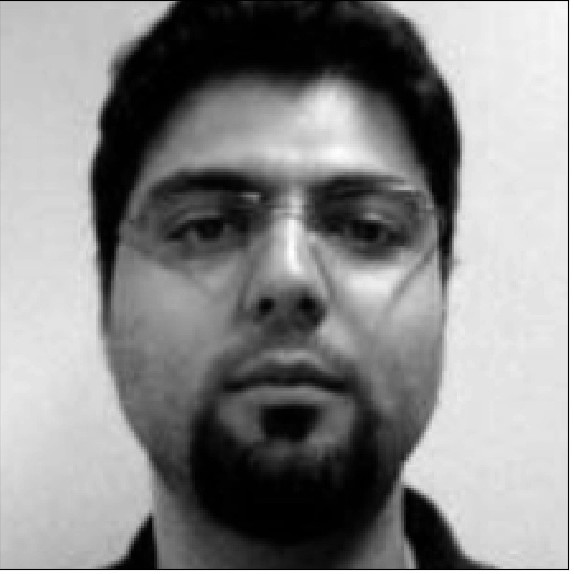}
        \includegraphics[width=.23\linewidth]{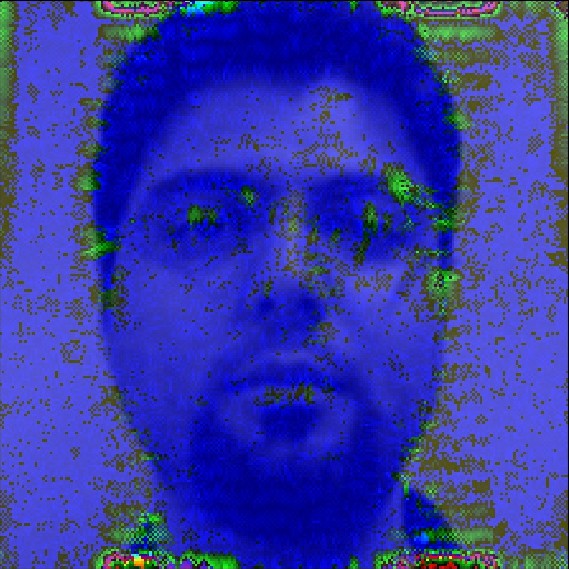}
        \includegraphics[width=.23\linewidth]{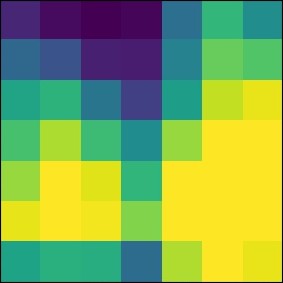}
        \includegraphics[width=.23\linewidth]{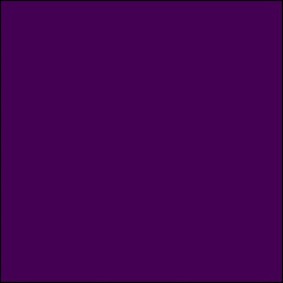}\\
        \includegraphics[width=.23\linewidth]{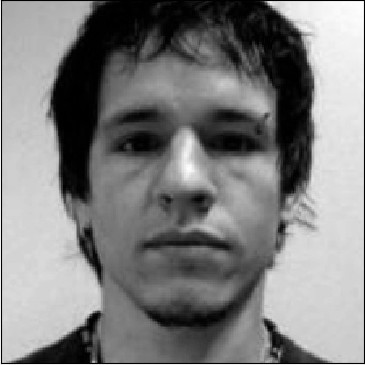}
        \includegraphics[width=.23\linewidth]{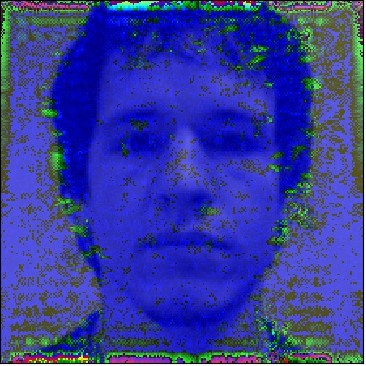}
        \includegraphics[width=.23\linewidth]{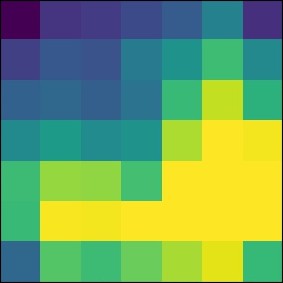}
        \includegraphics[width=.23\linewidth]{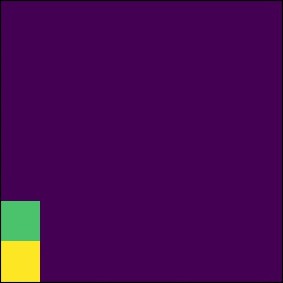}
        
        \caption{Bona-fide Face samples}
        \label{fig:freqRealFace}
    \end{subfigure}%
    \hfill
    \begin{subfigure}[t]{0.5\textwidth}
        \centering

        \includegraphics[width=.23\linewidth]{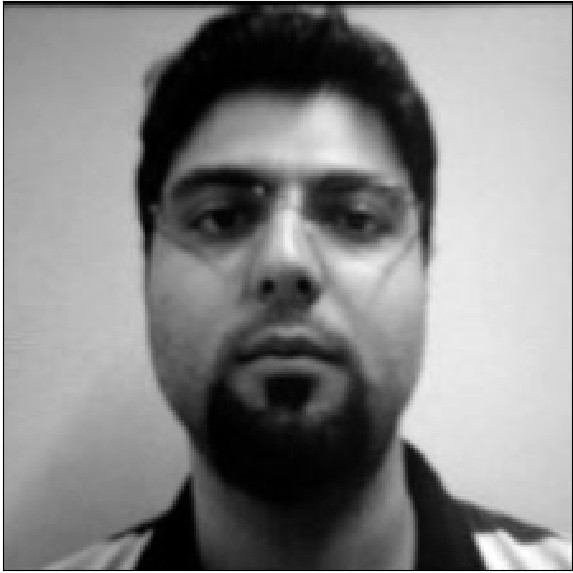}
        \includegraphics[width=.23\linewidth]{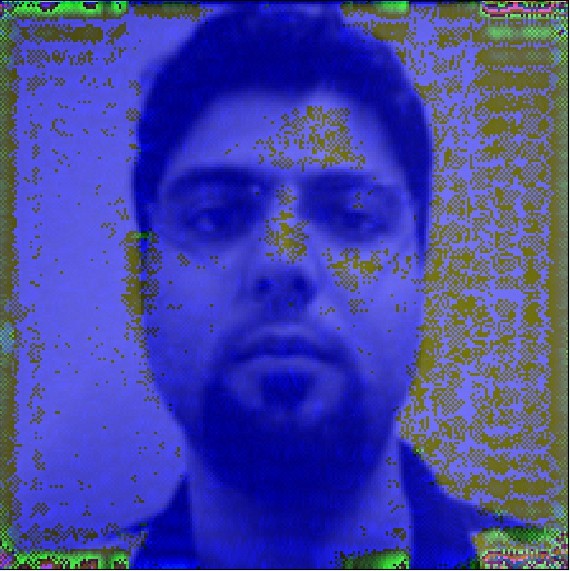}
		\includegraphics[width=.23\linewidth]{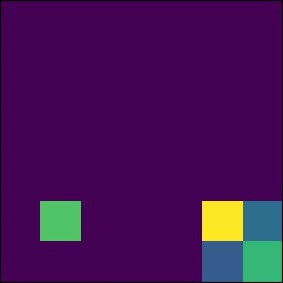}
		\includegraphics[width=.23\linewidth]{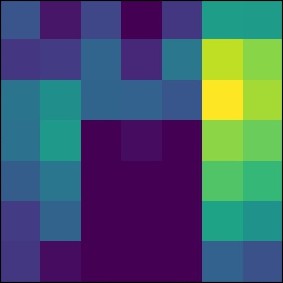} \\
        \includegraphics[width=.23\linewidth]{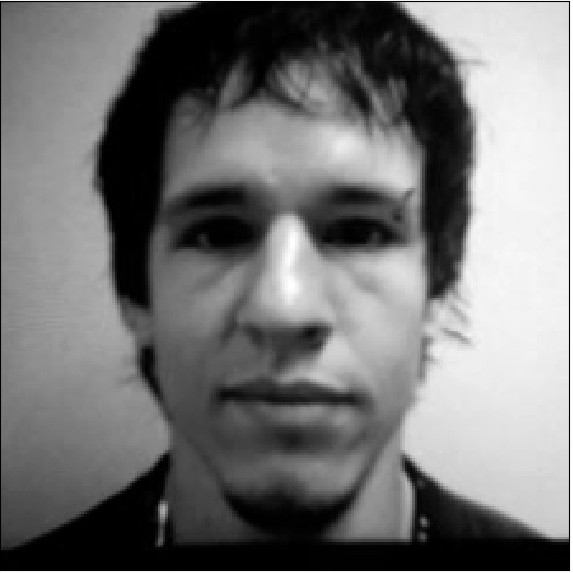}
        \includegraphics[width=.23\linewidth]{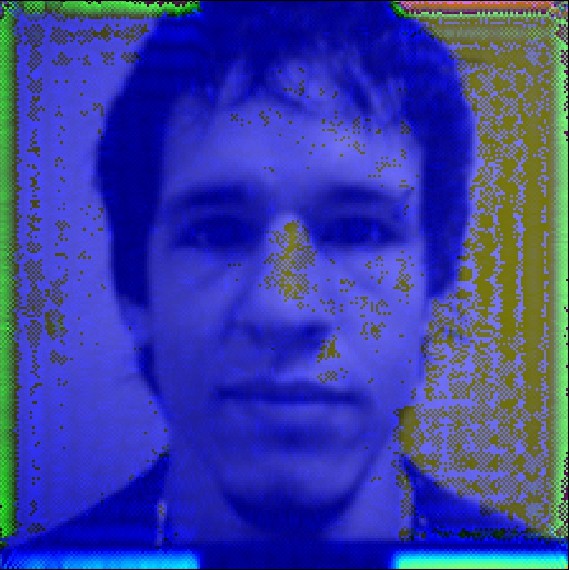}
        \includegraphics[width=.23\linewidth]{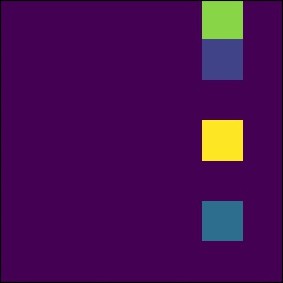}
        \includegraphics[width=.23\linewidth]{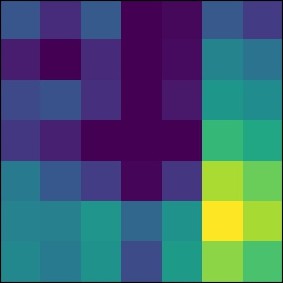}
        
        \caption{Attack Face samples}
        \label{fig:freqAttackFace}
    \end{subfigure}

    \caption{High frequency components in bona-fide vs Attack face images from Replay-Attack dataset.
     \figfooter{*}{Left-to-right: Input image, High frequency components overlayed over image, Network activation response of realF, Network activation response of attackF}
     \figfooter{*}{Top: client002, Bottom: client103}
}
    \label{fig:analysis_face}
\end{figure}

%-------------------------------------------------------------------------
%-------------------------------------------------------------------------

\begin{figure}[!] %[t!]
    \centering
    \begin{subfigure}[t]{0.5\textwidth}
        \centering
        \includegraphics[width=.15\linewidth]{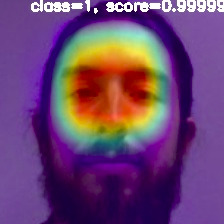}
        \includegraphics[width=.15\linewidth]{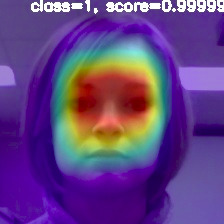}
        \includegraphics[width=.15\linewidth]{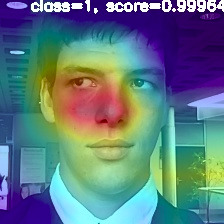}
        \includegraphics[width=.15\linewidth]{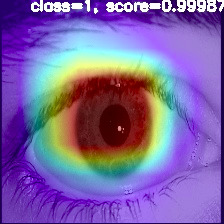}  
        \includegraphics[width=.15\linewidth]{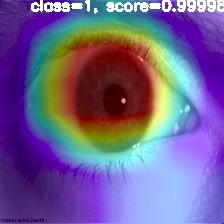}
        \includegraphics[width=.15\linewidth]{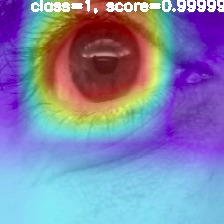}
        \caption{Correctly classified bona-fide images}
        \label{fig:correctReal}
    \end{subfigure}%
    \hfill
    \begin{subfigure}[t]{0.5\textwidth}
        \centering
        \includegraphics[width=.15\linewidth]{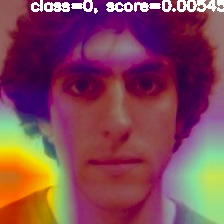}     
        \includegraphics[width=.15\linewidth]{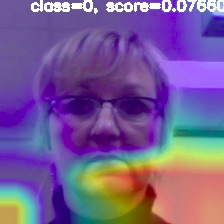}
        \includegraphics[width=.15\linewidth]{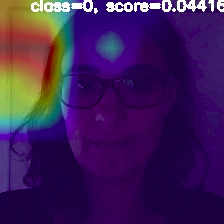}       
        \includegraphics[width=.15\linewidth]{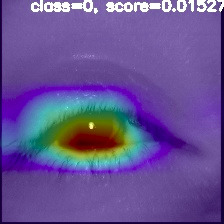} 
        \includegraphics[width=.15\linewidth]{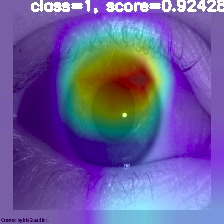}
        \includegraphics[width=.15\linewidth]{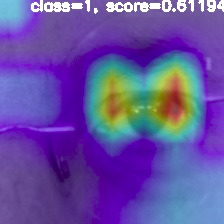}
        \caption{Incorrectly classified bona-fide images}
        \label{fig:falseREAL}
    \end{subfigure}

    \caption{Sample classified bona-fide images.
     \figfooter{*}{Left-to-right: Replay-Attack, MSU-MFSD - Replay-Mobile, ATVS-FIR , Warsaw, MobbioFake}
}
    \label{fig:gradcam_real}
\end{figure}

\begin{figure}[!]
    \centering
    \begin{subfigure}[t]{0.5\textwidth}
        \centering
        \includegraphics[width=.15\linewidth]{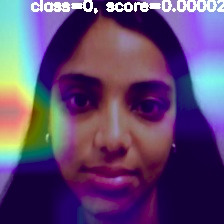}        
        \includegraphics[width=.15\linewidth]{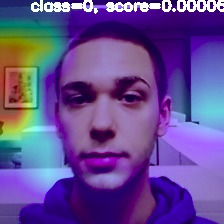}
        \includegraphics[width=.15\linewidth]{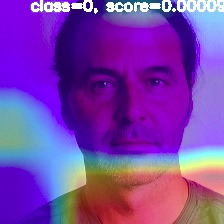}        
        \includegraphics[width=.15\linewidth]{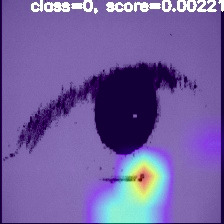}
        \includegraphics[width=.15\linewidth]{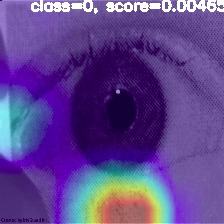}
        \includegraphics[width=.15\linewidth]{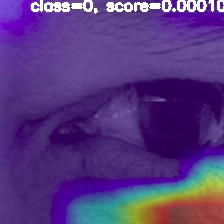}        
        \caption{Correctly classified Attack images}
        \label{fig:correctFake}
    \end{subfigure}%
    \hfill
    \begin{subfigure}[t]{0.5\textwidth}
        \centering
        \includegraphics[width=.15\linewidth]{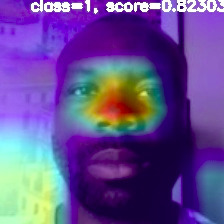}        
        \includegraphics[width=.15\linewidth]{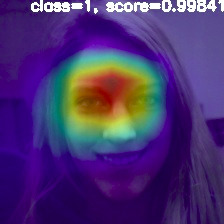}		\includegraphics[width=.15\linewidth]{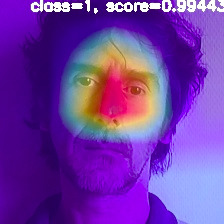}
        \includegraphics[width=.15\linewidth]{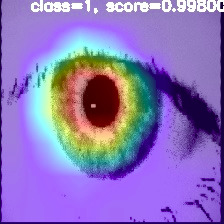} 
        \includegraphics[width=.15\linewidth]{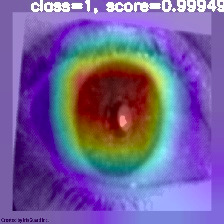}
        \includegraphics[width=.15\linewidth]{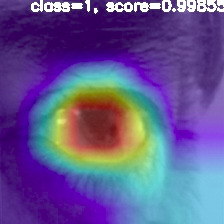}
        \caption{Incorrectly classified Attack images}
        \label{fig:falseFale}
    \end{subfigure}
    \caption{Sample classified Attack images.
     \figfooter{*}{Left-to-right: Replay-Attack, MSU-MFSD - Replay-Mobile, ATVS-FIR , Warsaw, MobbioFake}
}

    \label{fig:gradcam_fake}
\end{figure}

\subsection{Comparison with state of the art}

In Table~\ref{tabSOTA2} we compare our best achieved cross-datasets evaluation error with the state-of-the-art (SOTA).
This table shows that our cross-dataset testing results surpass SOTA on face datasets. For models trained on single biometric, InceptionV3 achieved best cross-dataset error on most datasets, except when training with MSU-MFSD, where MobilenetV2 achieved best test HTER on Replay-Attack and Replay-Mobile. While both models interchangeably achieved better error rate when network was trained using both face and iris images.

\subsection{Analysis and discussion}\label{sec:analysis}

Given results in Table~\ref{tabResFaceIris}, we feel that it is important to analyze features learnt by the networks that were trained on two biometrics. 

\subsubsection{Filters Activation and frequency components}

By training the network on several biometrics, we believe that the network is forced to learn to recognize specific patterns and texture in case of real presentations in general, which are not present in attack samples and vice-versa. An approach to prove that, is to visualize the response of certain feature maps from the network, given sample input images. For this task, we chose a MobilenetV2 network, trained with Iris-Warsaw and Face Replay-Attack datasets, from which we selected to visualize response of some filters from the final convolutional block before classification.

In addition to visualizing network activation response, we were also interested in viewing the distribution of high frequency components on the input images. We applied high pass filter to remove most of the low frequencies and keep only the high frequency components of an image, then we overlay these high frequency regions on the input image for better understanding of where high frequency is concentrated in bona-fide images vs attack images.

The steps for obtaining the regions with high frequency in image are depicted in Figure~\ref{fig:analysis_hpf}. Fourier transform was first applied to get frequency components in the image, High-pass-filter is then used to remove most of the low frequencies. After that, inverse Fourier is used to obtain an image with only the high frequency regions visible, and finally these regions are overlayed on original image to highlight those regions on the input image.

Regarding the chosen filters to visualize, the final layer before classification is consisted of 1280 filter, when an input image is of size $224 \times 224$, the feature maps resulting from these final 1280 filters have size $7 \times 7$. From these 1280 filters, we selected two filters for each biometric. One that had highest average activation response by real images, we call it \textit{realF}, and the other filter that was highly activated by attack images, called \textit{attackF}. We then visualize the $7 \times 7$ activation response of these filters for some bona-fide and attack samples belonging to the same subject.

Figure~\ref{fig:analysis_iris} shows samples from the iris Warsaw dataset with their high frequency regions highlighted in addition to activation response of \textit{realF} and \textit{attackF} features. The visualization clearly shows that for real iris samples, high frequency areas are concentrated at the bright regions around the inner eye corner and the lash-line, however, for attack presentations, high frequency regions mostly focus on noise patterns scattered on the attack medium (paper). The trained network could successfully recognize the real features highly represented at the eye corner, by \textit{realF} giving high response at those regions in real image, while failing to detect them in attack samples. \textit{AttackF} also manages to detect paper noise patterns in attack samples. which are missing in the real images.

Similar visualization for face images of Replay-Attack dataset is shown in Figure~\ref{fig:analysis_face}. The attack presentation in these samples is achieved using handheld mobile device showing a photo of the subject. We can notice the absence of high frequency response around the face edges, mouth and eyes, which are present in real samples and detected by the network through \textit{realF} feature. On the other hand, these attack images contain certain noise patterns on the clear background area which are detected successfully by the \textit{attackF} of the network.

By visualizing the high frequency regions in the images, it was proven that high frequency variations in case of real images, focus on certain edges and changes in the 3d real biometric presented. While in case of attack presentations, high frequency components represented noise, either spread on paper in case of paper attack medium, or noise patterns resulting from the display monitor of a mobile attack device. Our trained CNN network managed to correctly localize such features and lead to successful classification.

\subsubsection{Gradient-weighted class activation maps}

Another approach to analyze what the network learned, is by using gradient-weighted class activation maps~\cite{gradcam}; \textit{grad-cam}, which highlights the areas that contributed most to the network's classification decision for an input image.

We chose the trained MobilenetV2 network, on Replay-Attack face dataset and MobbioFake gray iris dataset, to analyze the activation maps for some bona-fide and attack test images from the 6 datasets. Figure~\ref{fig:gradcam_real} shows the \textit{grad-cam} for the bona-fide class given some real test samples that were either correctly of incorrectly classified, and Figure~\ref{fig:gradcam_fake} shows the same but for presentation attack images.

The visualizations in these figures show clearly that the network focuses on features of the face and iris centers to make its decision for a bona-fide presentation, and in cases when these center areas have lower activations, the network decides the image is an attack. And although the network was trained on two different biometrics, it could perfectly focus on the important center parts for each biometric individually.

\section{Conclusion and future work}\label{sec:conc}

In this paper, we surveyed the different face and iris presentation attack detection methods proposed in literature with focus on the use of very deep modern CNN architectures to tackle the PAD problem in both face and iris recognition. 

We compared two different networks; InceptionV3 and MobileNetV2, with two finetuning approaches starting from network weights that were trained on ImageNet dataset. Experimental results show that InceptionV3 achieved 0\% SOTA intra-dataset error on 6 publicly available benchmark datasets when finetuning the whole network's weights. The problem of small datasets sizes with very deep networks was overcome by augmenting the original data with random flipping, slight rotation and translation. In addition to starting the network training with non-random weights already pretrained on ImageNet dataset. This was to avoid the weights overfitting the small training dataset, however, for future work, we would like to check the effect of training these deep networks from scratch using the available PAD datasets.

The generalization ability of these networks has been tested by performing cross-dataset evaluation and better-than-SOTA results were reported for the face datasets. As far as we know this is the first paper to include cross-dataset Equal-Error-Rate evaluation for the three iris PAD datasets used, for which we achieved close to 0\% test EER. 

Not only cross-dataset evaluation was performed, but we also trained a single network using multiple biometric data; face and iris images. This commonly trained network achieved less cross-dataset error rates than networks trained with a single biometric type. This approach can be later applied on other biometric traits; such as the ear, to further demonstrate the effectiveness of using several biometric traits for training the PAD algorithm.

Finally, we provided analysis of the features learned by the networks, and showed their correlation with the frequency components present in input images regardless the biometric type. We also visualized the class heatmaps generated by the network for the bona-fide class which showed that the center of the face or iris are the most important parts that cause a network to select a bona-fide classification decision. For future work, we would like to apply patch-based CNN and identify the regions that contribute most to the attack detection in hope for improving the generalization of the approach by using a region-based CNN method.

\vfill\pagebreak

\bibliographystyle{iet}
\bibliography{y_IET-Submission-DoubleColumn-Template}

\end{document}